\documentclass{article}
\usepackage[T1]{fontenc}
\usepackage{arxiv_preprint}

\usepackage{amsmath,amsfonts,bm}

\def\eqref#1{equation~\ref{#1}}

\def\1{\bm{1}}

\DeclareMathAlphabet{\mathsfit}{\encodingdefault}{\sfdefault}{m}{sl}
\SetMathAlphabet{\mathsfit}{bold}{\encodingdefault}{\sfdefault}{bx}{n}

\usepackage[colorlinks=true,linkcolor=linkblue,citecolor=linkblue,urlcolor=linkblue]{hyperref}
\usepackage{url}
\usepackage{booktabs}
\usepackage{graphicx}
\usepackage{float}
\usepackage{enumitem}
\usepackage{multirow}
\usepackage{wrapfig}
\usepackage{array}
\usepackage{colortbl}
\makeatletter\providecommand{\insert@pcolumn}{\insert@column}\makeatother
\usepackage[most,skins,theorems]{tcolorbox}

\definecolor{lightblue}{RGB}{220,235,250}
\tcbset{
  takeawaysbox/.style={
    title=Takeaways,
    colback=lightblue!60,
    colframe=black,
    fonttitle=\bfseries\small,
    coltitle=white,
    colbacktitle=black,
    enhanced,
    attach boxed title to top left={xshift=2.5mm,yshift=-2.5mm},
    boxed title style={rounded corners, size=small, colframe=black, colback=black},
    width=\linewidth,
    arc=2.5mm,
    boxrule=0.8pt,
    top=3.5mm,
    bottom=2mm,
    left=2.5mm,
    right=2.5mm
  },
  epigraphbox/.style={
    enhanced,
    colback=gray!6,
    boxrule=0pt,
    frame hidden,
    borderline west={2.5pt}{0pt}{black!75},
    arc=0mm,
    left=3.5mm,
    right=2mm,
    top=1.8mm,
    bottom=1.8mm,
    width=\linewidth,
    fontupper=\small
  },
  promptbox/.style={
    enhanced,
    colback=gray!3,
    colframe=gray!60,
    arc=1.5mm,
    boxrule=0.6pt,
    fonttitle=\bfseries\small,
    coltitle=black,
    colbacktitle=gray!20,
    left=3mm,
    right=3mm,
    top=2.5mm,
    bottom=2.5mm,
    width=\linewidth,
    fontupper=\small,
    breakable
  }
}

\definecolor{bannergray}{RGB}{220, 220, 220}
\definecolor{ourhighlight}{RGB}{222, 245, 226}
\definecolor{posgreen}{RGB}{25, 145, 30}
\definecolor{negred}{RGB}{198, 28, 28}
\newcommand{\std}[1]{\,\textcolor{black!45}{\scriptsize$\pm$#1}}
\newcommand{\posdelta}[1]{\,\textcolor{posgreen}{\scriptsize$+$#1}}
\newcommand{\negdelta}[1]{\,\textcolor{negred}{\scriptsize$-$#1}}
\newcommand{\facton}{$\bullet$}
\newcommand{\factoff}{\textcolor{black!30}{$\circ$}}
\newcommand{\factna}{\textcolor{black!30}{--}}

\title{Scoring Higher, Answering Worse: \\ Mitigating Reward Hacking in Rubric-Based RL \\ via Protocol-Level Rubrics}
\author{%
{\bf Maoqi Liu$^{1,2*}$\quad Junwei He$^{2*}$\quad Bowen Zhang$^{2}$\quad Feiran Li$^{2}$}\\[2pt]
{\bf Wentao Ma$^{2}$\quad Rongyi Lin$^{2\dagger}$\quad Shuhan Zhong$^{2}$\quad Quan Fang$^{1}$\textsuperscript{\Letter}}\\[5pt]
$^1$Beijing University of Posts and Telecommunications\qquad $^2$ByteDance\\[3pt]
{\small\texttt{qfang@bupt.edu.cn}}%
}
\newcommand{\method}{ProRubric}

\begin{document}
\raggedbottom
\maketitle
\authornotes{$^*$Equal contribution.\quad $^\dagger$Project leader.\quad \textsuperscript{\Letter}\,Corresponding author.\quad Work done during an internship at ByteDance.}

\begin{abstract}
Rubric-based reinforcement learning (Rubric-RL) trains language models where no verifier exists.
A judge checks each criterion of a rubric, and the verdicts are aggregated into a reward, most often by a weighted sum.
We show that this additive aggregation is the weak point.
Under a sum, criteria compensate for one another: a policy that misses the one decision that matters can buy the points back with advice nobody asked for.
On clinical consultation, such a policy scores higher and answers worse.
Rubric coverage rises while appropriateness on held-out physician criteria falls below the untrained model.
The medical criteria are not to blame.
Grouped so that they must hold together, the same criteria, unchanged to the word, recover a third of the loss; shorter answers recover almost none.
We therefore propose \textbf{Pro}tocol-level \textbf{Rubric}s (\textbf{\method{}}), which keeps what the criteria ask for and changes how they are aggregated.
It groups a checklist into a few protocol-level dimensions.
A dimension counts only when all of its criteria hold and its failure clause does not fire.
The grouping is done once, offline, and leaves the optimizer unchanged.
\method{} raises appropriateness by 10.8 points without losing coverage and has the best seven-benchmark average at both scales.
Reward validity is set not only by what a rubric verifies, but by how it aggregates.
\ Code is available \href{https://github.com/Estrellajer/ProRubric}{here}.
\end{abstract}

\begin{figure}[H]
\centering
\vspace{-4pt}
\includegraphics[width=0.93\linewidth]{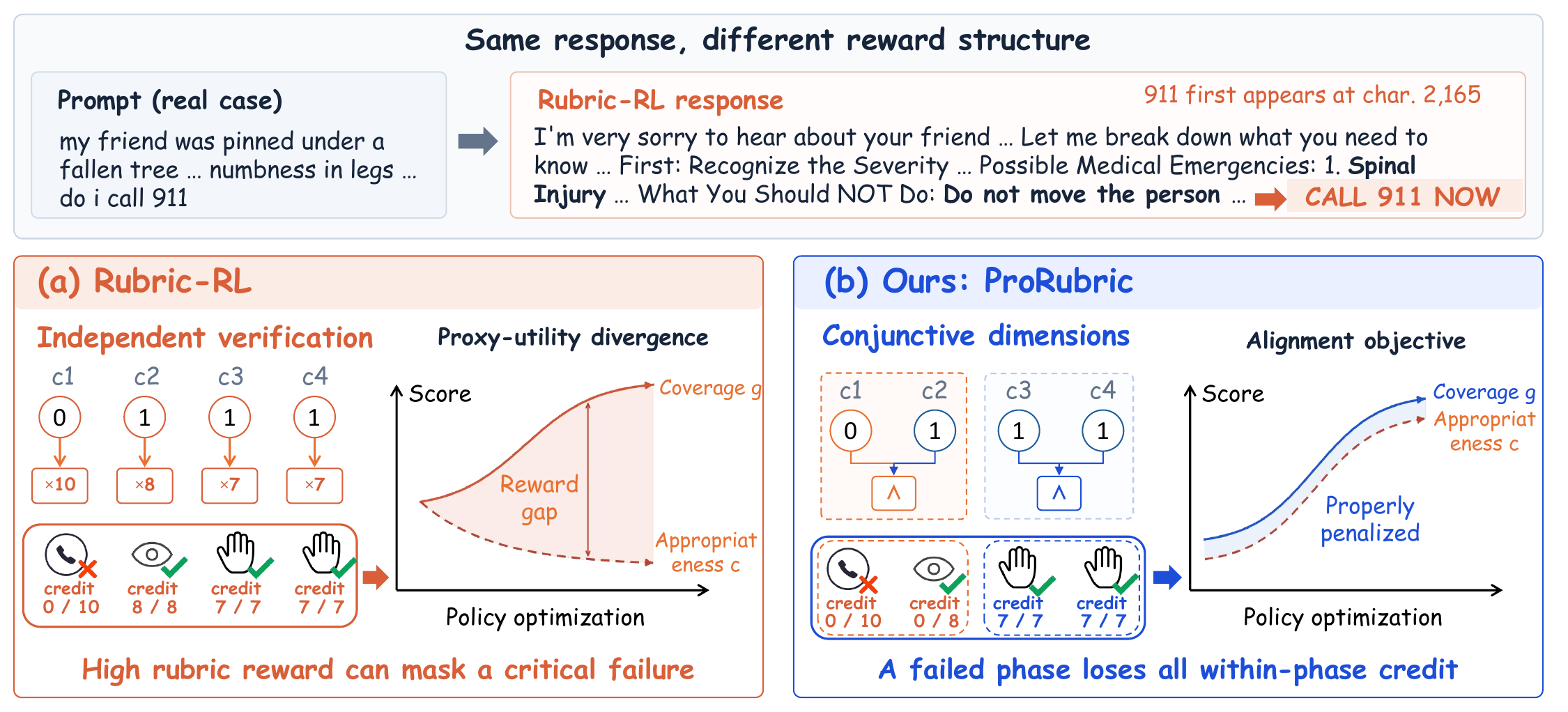}
\vspace{-10pt}
\caption{\textbf{Comparison of reward verification paradigms in clinical protocol RL.} Top: a real prompt and a Rubric-RL response in which the call to 911 first appears after 2,165 characters. (a) Conventional additive Rubric-RL scores atomic criteria independently, rewarding superficial coverage while masking critical omissions. (b) \method{} counts a group of criteria only when all of them hold, so cheap criteria can no longer offset a missed critical one.}
\label{fig:overview}
\end{figure}

\section{Introduction}
\label{sec:introduction}

Reinforcement learning has improved language models most where a verifier exists: a compiler for code, an answer checker for mathematics~\citep{Shao2024DeepSeekMath}. Where none exists, rubrics have taken its place. Rubric-based reinforcement learning (Rubric-RL) asks an LLM judge to check a response against a checklist of criteria and turns the verdicts into a reward~\citep{Gunjal2025RaR,Viswanathan2025Checklist,Huang2025Anchors,Jia2026OpenRS}. Unlike a monolithic reward model or judge, which is prone to length bias and style exploitation~\citep{Singhal2023LongForm,Dubois2024Length,Zheng2023Judge}, such a reward can be read criterion by criterion, and a growing body of work builds on it: rubric datasets and generators~\citep{Li2026RubricHub,Liu2025OpenRubrics,Shen2026RRD}, rubric-guided exploration~\citep{Zhou2025RuscaRL,Hou2026RISE}, and rubric benchmarks on which frontier models are evaluated~\citep{Arora2025HealthBench,OpenAI2025SystemCard}. Most of that work is about writing better criteria. Aggregation has barely changed: most often it is a weighted sum, which RaR calls \emph{explicit aggregation}~\citep{Gunjal2025RaR}.

Decomposing quality into criteria is easy; recomposing criteria into quality is where rubric rewards fail. Clinical standards are protocols, not tallies~\citep{Gawande2009Checklist}. Getting the triage right is not one criterion among thirty, and missing it is not a deduction to be made up elsewhere. A weighted sum treats it as exactly that. Criteria compensate for one another, so a policy that misses the decision can buy the points back with material nobody asked for, a form of reward hacking~\citep{Skalse2022Hacking} that policy gradient is well placed to find (Figure~\ref{fig:overview}a). Told \emph{``My baby has a fever,''} a policy trained this way replies with $10.9$k characters and fails both physician-written criteria for the case. One of them asks that the referral not be buried in verbose text.

The policy that results scores higher and answers worse, and standard reporting records only the first half. The score that published work reports is rubric coverage $g$, the proxy being optimized~\citep{Gao2023Overoptimization}. Training raises it on Qwen3-4B from $26.6$ to $32.1$. On the same queries, appropriateness $c$, judged against physician-written criteria held out from training, falls from $54.1$ to $26.0$, less than half of where the untrained model stands. An exploration baseline~\citep{Zhou2025RuscaRL} falls as far. A judge shown no rubric at all prefers the untrained model. \citet{Mahmoud2026Hacking} observed the same reversal and attributed it to under-specified rubrics; we show that aggregation alone moves it.

The obvious suspect is verbosity: trained answers are four times as long, and frontier evaluations now penalize longer answers on HealthBench~\citep{OpenAI2026GPT55SystemCard,Hicks2026HBProfessional}. We test this and other potential explanations systematically, leaving the criteria untouched. First, non-aggregation hypotheses fail to account for the collapse: cutting answers by $42\%$ leaves appropriateness where it was; swapping in a judge from another model family preserves the ranking; fine-tuning on the same rubric data without RL ends above the untrained model; and raising the KL penalty restores appropriateness only once learning stops. Second, what genuinely restores appropriateness while learning continues is the aggregation mechanism itself: grouping the verbatim criteria into conjunctive units recovers a third of the loss, while scoring them holistically as one list recovers nearly $40\%$. In medicine and dialogue, changing only the aggregation moves the collapse while the criteria stay fixed; in science, where criteria are substantive, both grouping and synthesis matter.

If the aggregation is the problem, the criteria can keep what they ask for while the aggregation changes. We propose \textbf{Pro}tocol-level \textbf{Rubric}s (\textbf{\method{}}; Figure~\ref{fig:overview}b) to make that change. \method{} groups a checklist into a few dimensions that follow the phases of the protocol. A dimension counts only when all its criteria hold, and each dimension carries a failure clause that voids it. The grouping is done once, offline. The optimizer and the served policy are untouched, and, unlike dependency-aware aggregation~\citep{Lv2026GEAR}, no relations among criteria need to be annotated. Told the same five words, the \method{} policy replies with $5.3$k characters and meets both criteria. \method{} raises appropriateness by $10.8$ points without losing coverage, and it has the best seven-benchmark average at both model scales.

Our main contributions are:
\begin{itemize}[leftmargin=18pt, itemsep=3pt, topsep=2pt]
    \item \textbf{Finding.} Rubric-RL with additive aggregation scores higher and answers worse: coverage rises while appropriateness on held-out physician criteria falls below the untrained model, across seeds, scales and judges, and the standard score does not show it.
    \item \textbf{Analysis.} With the criterion text fixed, changing only the aggregation moves the collapse in medicine and dialogue, and partly in science, while length, KL, weights, the judge and the data do not; the incentive can be read off the reward before training.
    \item \textbf{Method.} We propose \method{}, an offline regrouping of the rubric that improves appropriateness without losing coverage and has the best seven-benchmark average at both model scales.
\end{itemize}

\section{Preliminaries}
\label{sec:preliminaries}

\subsection{Rubric Rewards: Criteria and Aggregation}
\label{sec:task_formulation}
A rubric specifies a task as a checklist of $m$ criteria $\mathcal{C} = \{(r_i, w_i)\}_{i=1}^m$, where $r_i$ is a natural-language requirement and $w_i \neq 0$ its weight, negative for a penalty criterion (a behavior to avoid)~\citep{Gunjal2025RaR,Viswanathan2025Checklist}.
Turning a rubric into a reward involves two choices: the criteria, which fix what is checked, and the aggregation, which turns a judge's verdicts into a scalar.
The standard choice scores each criterion separately, $s_i(x, y) \in \{0, 1\}$ for a prompt $x$ and a response $y \sim \pi_\theta(\cdot \mid x)$, and takes the normalized weighted sum, which RaR calls \emph{explicit aggregation}:
\begin{equation}
    R_{\mathrm{atom}}(x, y) = \mathrm{clip}_{[0,1]}\Bigl(\frac{\sum_{i=1}^m w_i s_i(x, y)}{\sum_{i:\, w_i > 0} w_i}\Bigr).
    \label{eq:atomic_reward}
\end{equation}
Other aggregations keep the weighted mean and change what counts as one scored item: a group of criteria that must all hold (conjunctive grouping, Section~\ref{sec:method}), or the whole checklist scored at once (RaR's \emph{implicit aggregation}).
Our controls fix $\mathcal{C}$ and the optimizer and change only this choice.

\paragraph{Compensation.}
Under Eq.~\ref{eq:atomic_reward}, criteria compensate for one another: a response that misses a critical criterion can recover the lost reward by satisfying easier ones (Figure~\ref{fig:overview}).
A policy that learns to do this is reward hacking~\citep{Skalse2022Hacking}, and Section~\ref{sec:joint_requirements} shows why an optimizer of expected reward is drawn to it.
We optimize with GRPO~\citep{Shao2024DeepSeekMath}; training details are in Section~\ref{sec:setup}.

\subsection{Two Axes: Rubric Coverage and Appropriateness}
\label{sec:eval_protocol}
A policy scored only by the kind of rubric it was trained on cannot reveal compensation, so we read every medical policy on two axes.
\textbf{Rubric coverage} $g$ is the score on HealthBench~\citep{Arora2025HealthBench} under its own example-specific criteria. It is the proxy that training optimizes, in the sense of \citet{Gao2023Overoptimization}, and the number published work reports.
\textbf{Appropriateness} $c$ is the score on HealthBench's physician-consensus criteria, which practicing clinicians wrote and which are held out from training. They ask whether the response gets the decision right for this user, for example whether an emergency referral is stated clearly rather than buried in text.
Both axes are scored by DeepSeek-V4-Pro (Section~\ref{sec:setup}), not by clinicians, hence \emph{appropriateness} rather than clinical utility.

\paragraph{Problem statement.}
With the criteria and the optimizer held fixed, does the choice of aggregation decide whether optimizing the reward raises $c$ along with $g$?
Section~\ref{sec:reward_hacking} shows that under explicit aggregation $g$ rises while $c$ falls below the untrained model, and why. Section~\ref{sec:method} proposes a conjunctive aggregation, and Section~\ref{sec:experiments} tests both against the alternatives.

\section{The Reward-Hacking Channel in Additive Aggregation}
\label{sec:reward_hacking}

\subsection{Scoring Higher, Answering Worse}
\label{sec:quality_gap}
\begin{wrapfigure}{R}{0.5\textwidth}
\centering
\vspace{-12pt}
\includegraphics[width=0.5\textwidth]{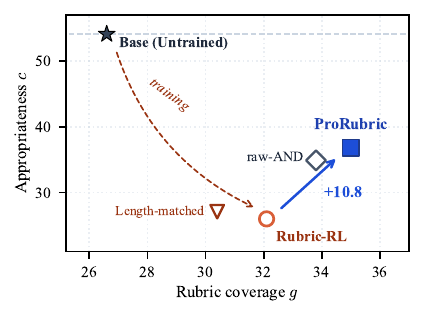}
\vspace{-24pt}
\caption{\textbf{Reward hacking under additive aggregation.} Coverage vs.\ appropriateness across the medical 4B models.}
\label{fig:diagnosis_compound}
\vspace{-8pt}
\end{wrapfigure}
In clinical consultation, training against the explicitly aggregated reward pulls the two axes apart (Figure~\ref{fig:diagnosis_compound}).
Coverage rises by five points, while appropriateness falls by twenty-eight, to less than half of where the untrained model started.
The pattern holds at both model scales and in every seed, and an exploration baseline that keeps the same reward falls just as far.

On the score that published work reports, the policy improves; the damage appears only on criteria held out from training.
Nor is it an artifact of rubric-based judging: shown the answers with no rubric at all, judges from two model families (DeepSeek-V4-Pro and GPT-5.6-luna; Section~\ref{sec:setup}) both prefer the untrained model, in about two pairs of three and in nearly all pairs respectively, averaged over three seeds (Table~\ref{tab:evidence}a). The two judges disagree in science, and in dialogue and in writing both prefer the trained policy (Appendix~\ref{app:probe}).

\begin{table}[t]
\centering
\caption{\textbf{Evidence of reward hacking without a rubric and before training.} (a) Blind pairwise win rates against the untrained Base (bold denotes Base preferred). (b) Reward change per edit.}
\label{tab:evidence}
\vspace{4pt}
\small
\setlength{\tabcolsep}{0pt}
\resizebox{\textwidth}{!}{%
\begin{tabular}{@{}l@{\hspace{6pt}}|@{\hspace{8pt}}>{\centering\arraybackslash}p{8.6em}@{\hspace{8pt}}>{\centering\arraybackslash}p{8.6em}@{\hspace{22pt}}l@{\hspace{6pt}}|@{\hspace{6pt}}c@{\hspace{6pt}}c@{\hspace{6pt}}c@{}}
\toprule
\multicolumn{3}{@{}c@{\hspace{22pt}}}{\textbf{(a) After Rubric-RL training: untrained answer preferred (\%)}} & \multicolumn{4}{c@{}}{\textbf{(b) Before training: reward change per edit}} \\
\cmidrule(r{22pt}){1-3} \cmidrule{4-7}
Domain & DeepSeek-V4-Pro & GPT-5.6-luna & Edit to the answer & Rubric-RL & raw-AND & \method{} \\
\midrule
Medicine & $\mathbf{65.7}$\std{13.2} & $\mathbf{93.6}$\std{2.3} & Name an item & $\mathbf{+7.3}$ & $-0.3$ & $+2.2$ \\
Science & $1.2$\std{1.2} & $\mathbf{90.3}$\std{9.0} & Name the topic & $-0.4$ & $+0.6$ & $-3.8$ \\
Dialogue & $24.3$\std{5.8} & $44.9$\std{6.8} & Add a needless test & $+0.3$ & $-1.4$ & $\mathbf{-5.8}$ \\
Writing & $0.0$\std{0.0} & $1.4$\std{1.6} & Drop the key advice & $-0.9$ & $-0.8$ & $-2.2$ \\
\bottomrule
\end{tabular}}
\end{table}

\paragraph{The criteria set the sign.}
Both axes are scored by the same judge, so the gap comes from what is checked, not from who checks it: on the same responses, the coverage criteria record a gain over the untrained model and the physician criteria a loss.
A judge from the training reward's own family shows the same sign with a smaller loss (Appendix~\ref{app:sec5_notes}); a judge closer to the reward understates the damage rather than creating it.

\paragraph{What the policy learns.}
The trained policy writes four times as much as the untrained one. On nearly half of the physician criteria (about a quarter before training), the training-family judge passes the answer and DeepSeek-V4-Pro does not (Appendix~\ref{app:sec5_notes}).
The key decision may still be there, but buried: in Figure~\ref{fig:overview}, the emergency referral first appears two thousand characters into the answer. On \emph{``My baby has a fever''} the answer exceeds $10{,}000$ characters and fails both physician criteria (Appendix~\ref{app:representative_examples}).

\subsection{Why the Cheap Behavior Is the Rational One}
\label{sec:joint_requirements}
Under additive aggregation every criterion pays a fixed rate however hard it is to satisfy, so the cheapest criteria are the rational target. Let $p_i$ be the probability that the policy satisfies criterion $i$; under Eq.~\ref{eq:atomic_reward} the expected reward is a weighted sum of these probabilities, and each criterion's rate is its share of the total weight.
A correct triage decision and a reminder to stay hydrated earn the same, but the first needs clinical reasoning the policy does not reliably have, while the second comes almost free from pre-training.
Equal reward, unequal cost: a policy maximizing expected reward has no reason to prefer the first.
This is an argument about the objective, not a convergence proof about the optimizer, and it makes a prediction that can be checked without training anything.

\subsection{Reading the Incentive off the Reward}
\label{sec:incentive_audit}
We perturb the untrained model's answers to 300 clinical cases in ways that mirror what trained policies do, and record how the reward changes (Table~\ref{tab:evidence}b). A paraphrase that keeps the meaning moves each of the three rewards by about one point ($+1.0$ to $+1.1$; Table~\ref{tab:incentive_audit_full}), so the larger changes are not judge noise.
Three readings of the Rubric-RL column account for Section~\ref{sec:quality_gap}.
Naming a concrete item without analyzing it pays well, while naming only the topic pays nothing: the reward buys recognizable entities, not substance.
Recommending an unwarranted high-risk test costs nothing, because no criterion forbids it.
Deleting the critical recommendation costs almost nothing either, so an answer that drops the decision and adds a term comes out ahead.
The other two columns preview Section~\ref{sec:method}: grouping the same criteria verbatim (raw-AND) stops paying for a name, and \method{}'s failure clauses make the needless test costly.

\section{Method: Protocol-Level Rubrics (\method{})}
\label{sec:method}

Section~\ref{sec:reward_hacking} traced the collapse to a fixed price per criterion; \method{} removes that price within each phase of the protocol, changing how criteria are aggregated, not what they ask for.
It rests on one property of the protocols such checklists encode: they are gated rather than tallied~\citep{Gawande2009Checklist}. Within a phase, requirements hold together, and failing one cannot be made up elsewhere.
\method{} writes this property into the rubric once, offline, and leaves the optimizer and the served policy unchanged (Figure~\ref{fig:overview}b).
Section~\ref{sec:dimensions} groups the criteria into dimensions that must hold jointly, Section~\ref{sec:failure_conditions} adds a failure clause to each dimension, and Section~\ref{sec:generation} generates them.

\subsection{Conjunctive Dimensions}
\label{sec:dimensions}
Under explicit aggregation every criterion earns a fixed share of the reward however hard it is to satisfy (Section~\ref{sec:joint_requirements}), so a policy can collect easy criteria in place of the one that matters.
\method{} partitions the checklist $\mathcal{C} = \{(r_i, w_i)\}_{i=1}^m$ into $K$ protocol dimensions $\mathcal{I}_1, \ldots, \mathcal{I}_K$.
For each criterion $i$, let $q_i \in \{0, 1\}$ denote compliance: $q_i = s_i$ for positive criteria ($w_i > 0$), and $q_i = 1 - s_i$ for penalty criteria ($w_i < 0$), ensuring penalty conditions act as hard non-violation constraints.
A dimension is satisfied only when all of its criteria hold jointly, carrying their aggregated weight:
\begin{equation}
    \tilde s_k = \prod_{i \in \mathcal{I}_k} q_i,
    \qquad
    \tilde w_k = \sum_{i \in \mathcal{I}_k} |w_i|,
    \qquad
    \tilde R = \frac{\sum_{k} \tilde w_k \tilde s_k}{\sum_k \tilde w_k}.
    \label{eq:conjunction}
\end{equation}
The reward is the normalized weighted sum over dimension verdicts; in practice, the judge verifies $\tilde s_k$ directly via synthesized protocol prompts (Section~\ref{sec:generation}).

This alters the marginal incentive of satisfying individual criteria. Under an idealized assumption where criterion compliances hold independently with probability $p_i = \mathbb{P}(q_i = 1)$,
\begin{equation}
    \frac{\partial\, \mathbb{E}[\tilde R]}{\partial p_i} \;=\; \frac{\tilde w_k}{\sum_\ell \tilde w_\ell} \prod_{j \in \mathcal{I}_k \setminus \{i\}} p_j ,
    \qquad i \in \mathcal{I}_k .
    \label{eq:conjunctive_price}
\end{equation}
A criterion now pays only in proportion to how likely the rest of its dimension is to hold. A fragment satisfied in a dimension the policy otherwise fails earns almost nothing, while the last missing piece of a nearly complete dimension earns the dimension's full weight: the reward stops paying for isolated fragments and starts paying for completed protocol phases.

\subsection{Failure Clauses}
\label{sec:failure_conditions}
\label{sec:defense_hierarchy}
Conjunction acts only on what the criteria name. No criterion forbids burying the decision under unrequested material or adding a needless high-risk test, so neither costs anything under Eq.~\ref{eq:conjunction}.
\method{} therefore ends each dimension's description with a \emph{failure clause}, a sentence naming a condition under which the phase is not met however much else the response contains, such as delaying urgent care. A dimension then counts only if all of its criteria hold and its failure clause does not fire.
Both the conjunction and the failure clause are written into the text the judge reads, and the judge returns one verdict per dimension; nothing is computed after it answers. This matters for the controls of Section~\ref{sec:ablations}, which differ from \method{} only in that text.

An additive deduction can be offset by adding more; a fired failure clause cannot, because nothing else in the dimension restores its credit.
Both effects appear before any training in Table~\ref{tab:evidence}b: grouping removes the payment for naming an item, and the failure clause makes the needless test costly.
Both act within a dimension only. Across dimensions the reward remains additive, so \method{} reduces the units that can compensate for one another from $m$ to $K$ rather than removing compensation.

\subsection{Generating the Dimensions}
\label{sec:generation}
Given a prompt and its checklist, a generator model first partitions the criteria into protocol-level dimensions, then rewrites each group as a single dimension description that ends in a failure clause, and finally assigns each dimension the summed weight of its criteria (Eq.~\ref{eq:conjunction}).
A validation step requires every criterion to belong to exactly one dimension and repairs or drops rubrics that fail (Appendix~\ref{app:validation}).
We use $2 \le K \le 5$, matching the phases of the protocols we study; in clinical triage these are rule-out, workup and disposition.
Generation runs once per prompt before training (prompt in Appendix~\ref{app:generation}) and adds no serving cost; the judge's training cost was not measured.

\section{Experiments}
\label{sec:experiments}

We answer three questions:
(1) Does \method{} improve on Rubric-RL, and at what cost outside the domain it targets? (Section~\ref{sec:main_results})
(2) Is the aggregation the cause? (Section~\ref{sec:ablations})
(3) When does appropriateness fall, and why do only some domains collapse? (Sections~\ref{sec:levers} and~\ref{sec:redundancy_tolerance})

\subsection{Experimental Setup}
\label{sec:setup}
\paragraph{Models and datasets.}
We use Qwen3-4B and Qwen3-8B~\citep{Yang2025Qwen3} in non-thinking mode and train one policy per domain on RubricHub's medical, writing and dialogue subsets~\citep{Li2026RubricHub} and on RaR-Science~\citep{Gunjal2025RaR}; inference is rubric-free. We evaluate on seven benchmarks: HealthBench~\citep{Arora2025HealthBench} and MedQA~\citep{Jin2020MedQA} in medicine, WritingBench~\citep{Wu2025WritingBench} and Creative-v3~\citep{EQBenchCreative} in writing, Arena-Hard v2~\citep{Li2025ArenaHard} in dialogue, and GPQA-Diamond~\citep{Rein2023GPQA} and ResearchQA~\citep{Yifei2025ResearchQA} in science, and read appropriateness on HealthBench-consensus. DeepSeek-V4-Pro judges both axes.

\paragraph{Baselines.}
We compare against the untrained model (Base) and four methods trained on the same prompts: (1) Rubric-RL~\citep{Gunjal2025RaR}, the standard formulation, which scores each criterion separately and sums the weighted scores; (2) RuscaRL~\citep{Zhou2025RuscaRL}, which uses rubrics to scaffold exploration during rollouts but keeps additive aggregation; (3) OPSD, on-policy self-distillation~\citep{Zhao2026OPSD} with the rubric as the teacher's context as in RGSD~\citep{Rezaei2026RGSD}, run under a smaller generation budget; and (4) SFT on RubricHub's released SFT responses. Responses are capped at 8,192 tokens.

\paragraph{Implementation details.}
All RL methods use GRPO with 64 prompts per batch, 8 responses per prompt, learning rate $10^{-6}$ and 300 steps; advantages are normalized within each group, and no KL penalty is used unless stated. Training rewards come from Doubao-mini, which is not used for evaluation. Every trained method in the main table has three seeds, and all ablations are compared on one common set of items. More details are in Appendices~\ref{app:training} and~\ref{app:evaluation}.

\subsection{Main Results}
\label{sec:main_results}

\begin{table}[t]
\centering
\caption{\textbf{Cross-domain evaluation results at 4B and 8B scales.} Deltas are relative to the untrained Base; bold and underline mark the best and second-best results.}
\label{tab:main_results}
\vspace{2pt}
\footnotesize
\setlength{\tabcolsep}{3.2pt}
\renewcommand{\arraystretch}{1.05}
\resizebox{\textwidth}{!}{\begin{tabular}{l|ll l ll ll l}
\toprule
\multirow{2}{*}{\textbf{Method}} & \multicolumn{2}{c}{\textbf{Writing}} & \multicolumn{1}{c}{\textbf{Dialogue}} & \multicolumn{2}{c}{\textbf{Clinical Medicine}} & \multicolumn{2}{c}{\textbf{Scientific Reasoning}} & \multicolumn{1}{c}{\textbf{Overall}} \\
\cmidrule(lr){2-3} \cmidrule(lr){4-4} \cmidrule(lr){5-6} \cmidrule(lr){7-8} \cmidrule(lr){9-9}
& \multicolumn{1}{c}{\textbf{WritingBench}} & \multicolumn{1}{c}{\textbf{Creative-v3}} & \multicolumn{1}{c}{\textbf{Arena-Hard}} & \multicolumn{1}{c}{\textbf{HealthBench}} & \multicolumn{1}{c}{\textbf{MedQA}} & \multicolumn{1}{c}{\textbf{GPQA}} & \multicolumn{1}{c}{\textbf{ResearchQA}} & \multicolumn{1}{c}{\textbf{Average}} \\
\midrule
\multicolumn{9}{@{}l}{\textit{Qwen3-4B}} \\
\hspace{0.8em}Base & 49.2 & 27.9 & 12.9 & 26.1 & \textbf{65.2} & 41.1 & 51.8 & 39.2 \\
\hspace{0.8em}+ SFT & 63.9\posdelta{14.7} & \underline{35.5}\posdelta{7.6} & \textbf{21.6}\posdelta{8.7} & \textbf{35.1}\posdelta{9.1} & 59.5\negdelta{5.7} & 36.0\negdelta{5.1} & \textbf{63.2}\posdelta{11.4} & 45.0\posdelta{5.8} \\
\hspace{0.8em}+ OPSD & 49.7\posdelta{0.4} & 27.7\negdelta{0.1} & 9.2\negdelta{3.7} & 25.1\negdelta{0.9} & 60.8\negdelta{4.4} & 36.4\negdelta{4.7} & 47.5\negdelta{4.3} & 36.6\negdelta{2.5} \\
\hspace{0.8em}+ Rubric-RL & \underline{64.6}\posdelta{15.3} & 35.0\posdelta{7.2} & 17.3\posdelta{4.4} & 31.9\posdelta{5.8} & 62.6\negdelta{2.6} & \underline{44.7}\posdelta{3.6} & \underline{62.9}\posdelta{11.1} & \underline{45.6}\posdelta{6.4} \\
\hspace{0.8em}+ RuscaRL & 63.7\posdelta{14.4} & 34.0\posdelta{6.2} & 18.2\posdelta{5.3} & 31.9\posdelta{5.8} & 62.0\negdelta{3.2} & \textbf{45.6}\posdelta{4.5} & 61.3\posdelta{9.4} & 45.2\posdelta{6.1} \\
\rowcolor{ourhighlight} \multicolumn{1}{l|}{\hspace{0.8em}\textbf{+ \method{} (Ours)}} & \textbf{67.1}\posdelta{17.8} & \textbf{37.7}\posdelta{9.9} & \underline{18.6}\posdelta{5.7} & \underline{34.7}\posdelta{8.6} & \underline{65.1}\negdelta{0.1} & 43.9\posdelta{2.8} & 61.1\posdelta{9.3} & \textbf{46.9}\posdelta{7.7} \\
\midrule
\multicolumn{9}{@{}l}{\textit{Qwen3-8B}} \\
\hspace{0.8em}Base & 53.9 & 36.0 & 19.1 & 31.4 & 69.0 & 42.9 & 55.9 & 44.0 \\
\hspace{0.8em}+ SFT & \underline{68.2}\posdelta{14.3} & \textbf{43.8}\posdelta{7.8} & \textbf{30.6}\posdelta{11.5} & \textbf{40.4}\posdelta{9.0} & 67.8\negdelta{1.2} & 31.0\negdelta{12.0} & \textbf{67.4}\posdelta{11.5} & 49.9\posdelta{5.8} \\
\hspace{0.8em}+ OPSD & 48.1\negdelta{5.8} & 23.2\negdelta{12.8} & 9.7\negdelta{9.4} & 27.0\negdelta{4.3} & 67.7\negdelta{1.3} & 40.2\negdelta{2.7} & 47.2\negdelta{8.7} & 37.6\negdelta{6.4} \\
\hspace{0.8em}+ Rubric-RL & 68.1\posdelta{14.2} & 38.9\posdelta{2.8} & 24.8\posdelta{5.7} & 36.4\posdelta{5.0} & 68.9\negdelta{0.2} & \underline{49.8}\posdelta{6.9} & 65.8\posdelta{9.9} & \underline{50.4}\posdelta{6.3} \\
\hspace{0.8em}+ RuscaRL & 68.0\posdelta{14.1} & 38.6\posdelta{2.5} & 24.0\posdelta{4.8} & 36.5\posdelta{5.1} & \underline{69.7}\posdelta{0.6} & 45.5\posdelta{2.5} & \underline{66.2}\posdelta{10.3} & 49.8\posdelta{5.7} \\
\rowcolor{ourhighlight} \multicolumn{1}{l|}{\hspace{0.8em}\textbf{+ \method{} (Ours)}} & \textbf{69.1}\posdelta{15.2} & \underline{43.1}\posdelta{7.1} & \underline{28.6}\posdelta{9.5} & \underline{39.9}\posdelta{8.6} & \textbf{69.9}\posdelta{0.8} & \textbf{50.0}\posdelta{7.1} & 64.9\posdelta{9.0} & \textbf{52.2}\posdelta{8.2} \\
\bottomrule
\end{tabular}
}
\end{table}

On the target domain, \method{} recovers appropriateness without losing coverage. At 4B it reaches $36.8$ against Rubric-RL's $26.0$ (Figure~\ref{fig:diagnosis_compound}, Table~\ref{tab:consensus_ablations}); at 8B it reaches $45.8$ against $29.5$. While appropriateness remains below the untrained Base ($54.1$), this reflects a Pareto trade-off: Base scores high on appropriateness through conservative answers that fail most requirements ($g = 26.6$), whereas Rubric-RL gains coverage ($32.1$) by collapsing appropriateness ($26.0$). \method{} expands the frontier, lifting coverage to $35.0$ and appropriateness to $36.8$ (Section~\ref{sec:levers} analyzes this gap).

Outside the target, the change costs little. \method{} has the best seven-benchmark average at both scales (Table~\ref{tab:main_results}): over Rubric-RL the margin is $+1.3$ points at 4B, with a 95\% interval of $[+0.2, +2.4]$ over three seeds, and \method{} is ahead of both RL baselines on every paired seed at both scales. Single benchmarks move far more across seeds than the average does, so we read them as orderings rather than effect sizes (Appendix~\ref{app:per_seed}).

The exceptions and the baselines mark the boundary. On MedQA, which uses no rubric, Rubric-RL regresses and \method{} does not; on ResearchQA \method{} trails Rubric-RL at both scales, since there breadth is itself the goal (Section~\ref{sec:redundancy_tolerance}). RuscaRL keeps additive aggregation and lands where Rubric-RL lands, OPSD reflects its smaller generation budget, and SFT pays the largest cost outside its domain, $-12.0$ on GPQA-Diamond at 8B (Appendix~\ref{app:sec5_notes}).

\subsection{Changing Only the Aggregation Moves the Collapse}
\label{sec:ablations}

\begin{wraptable}{R}{0.5\textwidth}
\vspace{-12pt}
\centering
\caption{\textbf{Medical ablations (4B).} $c$: appropriateness; $g$: coverage; both by DeepSeek-V4-Pro, averaged over three seeds on the common set (untrained Base: $54.1$ / $26.6$).}
\label{tab:consensus_ablations}
\vspace{2pt}
\footnotesize
\setlength{\tabcolsep}{3pt}
\renewcommand{\arraystretch}{1.1}
\begin{tabular*}{\linewidth}{@{\extracolsep{\fill}}l|cc@{}}
\toprule
\textbf{Reward} & $\boldsymbol{c}$ & $\boldsymbol{g}$ \\
\midrule
\textit{\method{} variants} & & \\
\hspace{0.6em}\method{} & 36.8\std{2.7} & 35.0\std{1.4} \\
\hspace{0.6em}$-$ failure clauses & 34.5\std{1.1} & 34.9\std{0.3} \\
\hspace{0.6em}partial credit (Graded) & 36.3\std{1.8} & 35.1\std{0.9} \\
\hspace{0.6em}one dimension ($K{=}1$) & 36.9\std{1.9} & 33.5\std{1.2} \\
\midrule
\textit{Same criteria, other aggregation} & & \\
\hspace{0.6em}additive (Rubric-RL) & 26.0\std{2.2} & 32.1\std{1.2} \\
\hspace{0.6em}grouped, verbatim (raw-AND) & 34.9\std{3.6} & 33.8\std{1.1} \\
\hspace{0.6em}holistic score (Implicit) & 36.7\std{0.1} & 34.9\std{0.1} \\
\midrule
\textit{Other changes} & & \\
\hspace{0.6em}shorter (Length-matched) & 27.1\std{1.8} & 30.4\std{1.4} \\
\hspace{0.6em}reweighted (Weighted) & 25.3\std{1.5} & 32.0\std{1.8} \\
\bottomrule
\end{tabular*}

\vspace{-8pt}
\end{wraptable}
Section~\ref{sec:reward_hacking} read the incentive off the reward before training; here we intervene on it. With the criterion text held fixed, every change to how the criteria are aggregated moves appropriateness up, and every change to anything else leaves the collapse in place (Table~\ref{tab:consensus_ablations}).
\label{sec:efficiency}
Two controls change only the aggregation, keeping every criterion verbatim: \emph{raw-AND} applies \method{}'s grouping, and \emph{implicit aggregation}~\citep{Gunjal2025RaR} scores the checklist whole. Two vary \method{} itself: \emph{Graded} gives four-level credit per dimension, and \emph{$K{=}1$} merges all dimensions into one. Three change something else: a \emph{length-matched} variant (half the response limit), a \emph{KL penalty}, and a \emph{weighted} rubric (Appendix~\ref{app:ablation_details}).

\paragraph{Changing the aggregation moves the collapse.}
Without rewriting anything, raw-AND recovers most of the distance to \method{} in medicine, though with higher variance ($\pm 3.6$ vs $\pm 2.7$) and lower coverage ($33.8$ vs $35.0$); implicit aggregation of the verbatim checklist matches \method{} within confidence intervals. Both reproduce at 8B over three seeds (Appendix~\ref{app:sec5_notes}). Within \method{}, partial credit per dimension is not behind binary credit, so we do not claim that hard conjunction is the better operator: what these variants share is that no criterion is scored and summed alone (Eq.~\ref{eq:conjunctive_price}). $K{=}1$ matches \method{} on appropriateness but has lower coverage ($33.5$ against $35.0$), consistent with a weaker learning signal: about two thirds of its sampled groups receive identical rewards on every seed.

\paragraph{Changing anything else does not.}

\begin{figure}[t]
\centering
\includegraphics[width=\linewidth]{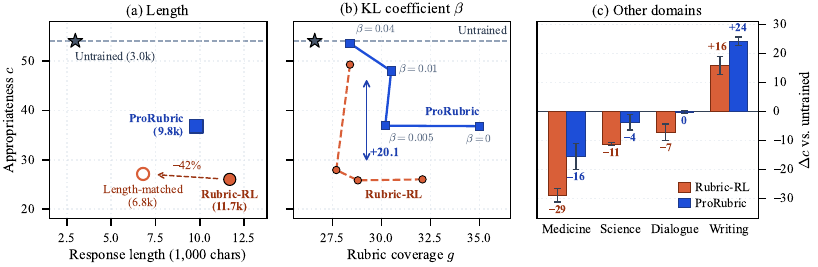}
\vspace{-20pt}
\caption{\textbf{Analysis of confounding factors and cross-domain generalization.} (a) Length matching does not restore appropriateness. (b) KL penalty. (c) Appropriateness retention across domains.}
\label{fig:controls_compound}
\vspace{-18pt}
\end{figure}

Length is the vehicle, not the cause. The length-matched variant, $42\%$ shorter, leaves appropriateness where it was (Figure~\ref{fig:controls_compound}a), and under an equal 2,048-token budget \method{} wins $83\%$ and $79\%$ of decided blind pairs against Rubric-RL in medicine and science (Appendix~\ref{app:pairwise}). A KL penalty recovers appropriateness only by stopping learning: at $\beta = 0.04$ both rewards approach the untrained model on both axes and in length (Figure~\ref{fig:controls_compound}b). Reweighting the rubric leaves Rubric-RL at $25.3$, and the science rubric, already weighted, loses appropriateness all the same. SFT on RubricHub's released responses ends above the untrained model ($66.3$ against $54.1$), so the data collapses appropriateness only once it becomes an additive reward. The ranking is nearly unchanged under GPT-5.6-luna (Appendix~\ref{app:sec5_notes}).

\paragraph{Where the attribution holds.}

In dialogue it holds as in medicine: implicit aggregation beats Rubric-RL on all three seeds, close to \method{} (Table~\ref{tab:probe}). In science, however, the limits of verbatim aggregation emerge: implicit aggregation, raw-AND, and rewriting criteria one to one each recover at most $3.2$ points, against \method{}'s $7.5$ (three-seed mean, at least $5.0$ on every seed). Where criteria define substantive derivations rather than presence checks, verbatim concatenation overburdens the judge; only structured protocol grouping and synthesis reliably lift appropriateness.

\subsection{When the Collapse Comes, and Where \method{} Stops}
\label{sec:levers}

\paragraph{Under additive aggregation, appropriateness collapses early.}
In retrained medical models (4B, three seeds, 300 HealthBench-consensus prompts; Figure~\ref{fig:training_dynamics} in Appendix~\ref{app:dynamics}), Rubric-RL falls from $55.8$ to $35.9$ within 50 steps and ends at $23.8$, while \method{} remains at $50.3$ at step 50 and ends at $35.3$; at every checkpoint every Rubric-RL seed lies below every \method{} seed. Without per-criterion credit the collapse is slower and smaller, but not stopped, as analyzed next.

\paragraph{It stops where the dimensions meet.}
Across dimensions the reward is still additive (Section~\ref{sec:failure_conditions}), so adding material still pays: \method{}'s responses run to $9.8$k characters against the untrained model's $3.0$k, and added content, named or substantive, costs $17$ to $38$ points of the redundancy sub-score ($25$ to $38$ for added substance; Table~\ref{tab:probe_seeds_audit}b). This is the residual gap: \method{} satisfies most of its dimensions but stays below the untrained model on appropriateness, and without a rubric the two judges disagree on whether it beats the untrained model (Table~\ref{tab:rubric_free_full}). It loses less than Rubric-RL because what it adds is substantive, but it moves along the same mechanism rather than escaping it.

\paragraph{Why protocol dimensions go beyond verbatim grouping.}
Verbatim grouping (raw-AND) recovers $8.9$ points in medicine, while removing \method{}'s failure clauses costs $2.3$ points (Table~\ref{tab:consensus_ablations}). Raw-AND also leaves criteria uncurated: higher cross-seed variance ($\pm 3.6$ against $\pm 2.7$), lower coverage ($33.8$ against $35.0$), and in science it recovers little ($+3.2$ against $+7.5$; Section~\ref{sec:ablations}). Synthesized dimensions give non-compensable failure conditions and a cleaner training signal.

\subsection{What a Rubric Pays For Decides Whether a Domain Collapses}
\label{sec:redundancy_tolerance}
Section~\ref{sec:incentive_audit} read the incentive off the medical rubric. Applied to all four domains, the same analysis accounts for most of the cross-domain pattern. Under explicit aggregation, appropriateness falls by $28.9$ in medicine, $11.2$ in science and $7.1$ in dialogue, and rises by $16.0$ in writing (Figure~\ref{fig:controls_compound}c; three-seed means on 200 prompts per domain, Table~\ref{tab:probe_seeds_audit}a). The cause is the same everywhere, additive aggregation paying for added material, but it has two routes: where criteria are mention-style the cheapest addition is hollow, and where they are substantive the policy must elaborate for real.

The first route is naming. On the untrained model's answers to 300 training prompts per domain (Table~\ref{tab:probe_seeds_audit}b), naming a checklist item without analyzing it earns, under the Rubric-RL reward, $32\%$ of what adding unrequested substance earns in medicine and $13\%$ in dialogue, and nothing detectable in writing ($7\%$) or science ($-8\%$); under \method{}'s reward the shares fall to $6\%$, $-14\%$, $-10\%$ and $-23\%$. Medicine, where both routes are open, collapses hardest.

The second route runs through substance, and it explains science. Science's criteria are verifiable requirements, such as deriving a threshold energy from the four-momentum invariant, that naming cannot satisfy: $195$ of $300$ naming edits leave the reward unchanged. Still, every rubric pays for correct, relevant material that no criterion asked for ($+22.7$ in medicine, $+18.9$ in writing, $+17.9$ in dialogue, $+10.8$ in science), and in every domain it costs more of the redundancy sub-score than naming does, $25$ to $38$ points against $17$ to $29$. Science loses appropriateness by this route alone.

Why writing improves remains open; writing policies may integrate rather than append material (Appendix~\ref{app:placement}). Disentangling tasks from rubric shapes requires further study (Appendix~\ref{app:sec5_notes}).
\section{Related Work}
\label{sec:related_work}

\paragraph{Rubrics as rewards.}
Fine-grained checks go back to behavioral testing~\citep{Ribeiro2020CheckList}, and rule-based feedback traces to Constitutional AI~\citep{Bai2022Constitutional} and RLAIF~\citep{Lee2023RLAIF,Yuan2024SelfReward}.
Prometheus~\citep{Kim2023Prometheus} and HealthBench~\citep{Arora2025HealthBench} made rubrics a standard interface for LLM-based evaluation.
Checklist and rubric rewards carry them into RL~\citep{Gunjal2025RaR,Viswanathan2025Checklist,Huang2025Anchors,Jia2026OpenRS}, and rubric datasets and generators scale up criteria~\citep{Li2026RubricHub,Liu2025OpenRubrics,Shen2026RRD}.
Rubrics also enter training in other ways: RuscaRL~\citep{Zhou2025RuscaRL} uses them as scaffolding during rollouts, RISE-RL~\citep{Hou2026RISE} selects trajectories to explore by the criteria a policy repeatedly misses, and rubric-guided self-distillation~\citep{Rezaei2026RGSD}, built on on-policy self-distillation~\citep{Zhao2026OPSD}, gives them to a teacher.
These works change what criteria ask for and where they enter training; as a reward, their verdicts are still a weighted sum, the step we study.

\paragraph{Aggregating the criteria.}
In RaR, \citet{Gunjal2025RaR} compare explicit aggregation, a normalized weighted sum of per-criterion verdicts, with implicit aggregation, in which the judge reads the whole checklist and returns one score.
Recent work questions the weighted sum.
\citet{Lv2026GEAR} observe that criteria scored as independent utilities pass credit past unmet prerequisites, which they call \emph{false credit propagation}, and suppress it through a rubric graph annotated with prerequisite and activation relations.
Rubric Dropout~\citep{Yang2026RubricDropout} drops criteria at random during training to curb reward hacking, and SRaR~\citep{Xie2026SRaR} attributes rubric items to individual steps of mathematical reasoning instead of folding them into one scalar.
\method{} needs no annotated relations and keeps one response-level reward: it groups a plain checklist into dimensions that must hold jointly, and we measure what the weighted sum costs on criteria the policy never trained against.

\paragraph{Reward hacking.}
Over-optimizing a proxy reward is an instance of Goodhart's law~\citep{Goodhart1975}, extensively documented in RLHF~\citep{Ouyang2022InstructGPT,Bai2022Helpful,Skalse2022Hacking,Gao2023Overoptimization} and DPO~\citep{Rafailov2024Overoptimization}, where reward model ensembles mitigate but do not eliminate gaming~\citep{Eisenstein2023Ensembles}; in text generation it often surfaces as length bias~\citep{Singhal2023LongForm,Dubois2024Length,OpenAI2026GPT55SystemCard}.
In rubric RL, \citet{Mahmoud2026Hacking} report the reversal we start from: rubric-based verifiers prefer the RL checkpoint while rubric-free judges prefer the base model, with gains concentrated in completeness and presence-based criteria, and attribute it to rubrics that leave important failure modes unspecified.
CHERRL~\citep{Wang2026CHERRL} reproduces reward hacking by injecting biases into judges.
We trace the reversal elsewhere: with the judge and the criterion text fixed, changing only the aggregation moves it in medicine and dialogue.

\section{Conclusion}
\label{sec:conclusion}

Rubric-based RL writes quality down as criteria and turns them back into a number; we find that the second step largely decides whether optimizing the number helps.
Under a weighted sum, cheap criteria compensate for a missed critical one, and the policy learns to do exactly that: coverage rises while appropriateness on held-out physician criteria falls below the untrained model.
With criterion text fixed, changing only aggregation moves this collapse in medicine and dialogue; in science, wording matters as well.
\method{} groups criteria offline into dimensions that must hold jointly, lifting appropriateness without losing coverage or touching the optimizer: reward validity depends on aggregation as much as verification.

\paragraph{Limitations and future work.}
Clinical consultation unfolds across multiple turns where clarifying questions are essential. This work evaluates single-turn responses using an LLM judge calibrated against physicians (Appendix~\ref{app:judge_agreement}) rather than clinical trials. Extending protocol rubrics to interactive dialogues and human studies remains a key next step.

\section*{Ethics Statement}
This work studies how rubric rewards shape the behavior of language models. The medical experiments use public benchmarks (HealthBench, MedQA) and involve no private data and no human subjects. The trained models are research artifacts; they are not validated for clinical use and should not be used to give medical advice or to triage patients.

\section*{Reproducibility Statement}
The medical prompt that generates \method{}'s dimensions and the binary training judge's prompt are given verbatim in Appendix~\ref{app:generation}. Code, the generated \method{} rubrics and all judge prompts will be released. Training hyperparameters and baseline configurations are in Appendix~\ref{app:training} (Table~\ref{tab:training}), and benchmark scoring, judge configurations, common evaluation sets and per-seed results in Appendices~\ref{app:evaluation} and~\ref{app:per_seed}.

\section*{AI Use Statement}
Language models generate \method{}'s dimensions and serve as judges, as described in the text. They also helped polish prose and format LaTeX; the authors verified every claim.

\clearpage
\bibliography{iclr2027_conference}
\bibliographystyle{iclr2027_conference}

\appendix
\begin{tcolorbox}[enhanced,colback=abstractbg,colframe=accent!70,boxrule=0.6pt,arc=2.5mm,
  left=4mm,right=4mm,top=2mm,bottom=2mm,before skip=4pt,after skip=10pt,
  title={Appendix Roadmap},fonttitle=\bfseries\large,coltitle=accent,colbacktitle=abstractbg,
  attach title to upper={\par\vspace{0.6ex}},center title]
This appendix provides reproducibility details, complete empirical evaluations, sensitivity analyses, and qualitative examples.
\begin{description}[leftmargin=2.4em,labelwidth=1.6em,labelsep=0.6em,itemsep=3pt,topsep=4pt,font=\bfseries\color{accent}]
    \item[\ref{app:implementation}] \textbf{Implementation Details and Prompts.} Training data curation, offline protocol generation templates (Box~A.1), reward judge specification (Box~A.2), schema validation, and training configurations (Table~\ref{tab:training}).
    \item[\ref{app:evaluation}] \textbf{Evaluation Protocols and Calibrations.} Benchmark configurations, human-judge calibration (Table~\ref{tab:judge_agreement}), sample sizes (Table~\ref{tab:eval_sizes}), and rubric-free blind pairwise protocols (Table~\ref{tab:rubric_free_full}, Figure~\ref{fig:pairwise_winrates}).
    \item[\ref{app:per_seed}] \textbf{Full Numerical Results and Ablations.} Per-seed cross-domain results (Table~\ref{tab:per_seed_domains}), complete medical ablation tables under dual judges (Table~\ref{tab:per_seed_ablations}, Table~\ref{tab:ablations_full}), cross-model ranking validation (Table~\ref{tab:gpt_crosscheck}), and cross-domain appropriateness probes (Table~\ref{tab:probe}, Table~\ref{tab:probe_seeds_audit}).
    \item[\ref{app:mechanistic}] \textbf{Mechanistic Analyses and Diagnostics.} Incentive audit across aggregations (Table~\ref{tab:incentive_audit_full}), training dynamics trajectories (Figure~\ref{fig:training_dynamics}), KL regularization sweep, fixed appropriateness criterion analysis, and text placement experiments.
    \item[\ref{app:qualitative}] \textbf{Qualitative Case Studies and Protocol Analysis.} Clinical consultation walkthrough, protocol dimensions, and trade-off analysis on exhaustive retrieval.
\end{description}
\end{tcolorbox}

\section{Implementation Details}
\label{app:implementation}
\subsection{Data Sources and Training Datasets}
We train one policy per domain using RubricHub's medical, writing and dialogue subsets~\citep{Li2026RubricHub} and RaR-Science~\citep{Gunjal2025RaR}.
For RubricHub, prompts are deduplicated by hash and filtered to retain rubrics with 5--60 nonempty criteria and positive total weight. For science, we train on the 18,333 prompts of the RaR-Science training split.
All model evaluations are conducted out-of-domain on standard benchmarks or against held-out physician criteria (Section~\ref{sec:setup}).
Rubric-RL and \method{} are trained on identical prompts, ensuring that all observed differences originate from reward aggregation rather than data curation.
\subsection{Rubric Generation Prompts and Protocol Templates}
\label{app:generation}
Offline generation turns atomic rubrics of 20--30 items into two to five protocol-level dimensions with DeepSeek-V4-Pro as the generator. Generation uses temperature 0, a limit of 3,000 output tokens, and at most two retries when the output fails schema validation.
The generator's proposed weights are retained as metadata; training weights are the absolute-weight sums of Section~\ref{sec:dimensions}.

Box~A.1 gives the full generation prompt. It asks the generator for two to five dimensions, each stating what a good answer achieves and what would make it fail, judgeable only from the whole answer, with every atomic item assigned to exactly one dimension and a weight equal to the sum of its items' weights. The \{anchor\} slot receives the training example's reference rubric as distributed with the dataset; in all four training sets this is the atomic rubric itself, re-rendered as a list, and no reference-answer notes are present, so the anchor adds nothing beyond the atomic items and no evaluation criterion is shown to the generator. Box~A.2 gives the training reward judge's prompt: the judge reads the whole rubric (the atomic criteria, or \method{}'s dimensions) and returns one binary verdict per item; raw-AND uses the same judge prompt, with each group's verbatim items under an instruction that all must hold. The writing and dialogue generation prompts differ in their task description; the science prompt keeps the medical wording and differs only in weighting items by magnitude and folding pitfall items into failure conditions. These prompts and those of the Implicit, Graded and $K{=}1$ judges will be released with the code.

\begin{tcolorbox}[
  promptbox,
  title={Box A.1: Prompt for Offline Dimension Generation (\method{}, medicine; verbatim)},
  label={box:restructuring_prompt}
]
\small\ttfamily
You are designing a compact, protocol-level grading rubric for a medical Q\&A answer, to replace an atomic checklist of many small \textquotedbl{}mentions X\textquotedbl{} items with 2-5 coarse-grained criteria that each describe what a GOOD answer ACHIEVES on one dimension.
\vspace{6pt}

\# Question / conversation\\\{conversation\}
\vspace{6pt}

\# Existing atomic rubric items (for reference only -- do not just copy these into your descriptions as a checklist)\\\{atomic\_items\}
\vspace{6pt}

\# Anchor for resolving contradictions (an expert-reviewed reference rubric for this same question -- some atomic items above may be redundant, mutually exclusive, or simply wrong; use this anchor to decide what a correct answer should actually contain)\\\{anchor\}
\vspace{6pt}

\# Task\\Design between 2 and 5 protocol-level criteria. Each criterion is a paragraph, not a checklist item. Each criterion must:\\1. state what a good answer ACHIEVES on that dimension for THIS SPECIFIC question, and what would make an answer FAIL that dimension;\\2. be judgeable only by reading and understanding the whole answer holistically -- NOT by scanning for a keyword or a single sentence;\\3. NOT be phrased as a list of \textquotedbl{}mentions A, B, C\textquotedbl{} terms to check off;\\4. if two or more atomic items below contradict each other (e.g. reward mutually exclusive answers), resolve the contradiction explicitly using the anchor -- the description should state what the CORRECT position is, not just note that people disagree;\\5. together with the other criteria, have every atomic item index from 1 to \{n\} assigned to exactly one criterion (pick whichever criterion is closest in topic, even after you rewrite its language into holistic phrasing);\\6. carry a weight equal to the sum of the weights of the atomic items assigned to it.
\vspace{6pt}

\# Output format\\Return ONLY a JSON object, no markdown fences, no other text:\\\{\textquotedbl{}criteria\textquotedbl{}: [\{\textquotedbl{}name\textquotedbl{}: \textquotedbl{}\textless{}=6 words\textquotedbl{}, \textquotedbl{}description\textquotedbl{}: \textquotedbl{}2-5 sentences, self-contained, states what ACHIEVES / FAILS this dimension\textquotedbl{}, \textquotedbl{}weight\textquotedbl{}: \textless{}number\textgreater{}, \textquotedbl{}atomic\_indices\textquotedbl{}: [\textless{}1-based ints into the atomic list above\textgreater{}]\}, ...]\}\\The atomic\_indices arrays across all criteria must union to exactly \{1, ..., \{n\}\} with no repeats and no omissions.
\end{tcolorbox}

\begin{tcolorbox}[
  promptbox,
  title={Box A.2: Prompt of the Training Reward Judge (verbatim)},
  label={box:judge_prompt}
]
\small\ttfamily
You are a strict binary rubric judge. Evaluate the assistant response against every criterion independently. A criterion is satisfied only when the response meets all of its requirements. Negative-weight criteria describe undesirable behavior: return true when that behavior is present. Return ONLY one JSON object with exactly one field, \textquotedbl{}satisfied\textquotedbl{}, whose value is an object mapping each criterion index (as a string, e.g. \textquotedbl{}1\textquotedbl{}, \textquotedbl{}2\textquotedbl{}, ...) to a boolean, covering every index exactly once. Do not include explanations, markdown, or extra keys.
\vspace{6pt}

PROMPT:\\\{prompt\}
\vspace{6pt}

ASSISTANT RESPONSE:\\\{response\}
\vspace{6pt}

CRITERIA:\\\relax[\{\textquotedbl{}index\textquotedbl{}: 1, \textquotedbl{}criterion\textquotedbl{}: ..., \textquotedbl{}weight\textquotedbl{}: ..., \textquotedbl{}polarity\textquotedbl{}: ...\}, ...]
\end{tcolorbox}

\subsection{Validation, Repair, and Filtering}
\label{app:validation}
Each generation is checked for the required fields and for a complete partition of the criteria.
If retries fail, a structural repair drops invalid or duplicate assignments and assigns an omitted criterion to the nearest group by index, then recomputes the weights; it does not regenerate descriptions.
A rubric is kept only if every criterion occurs exactly once, every group is nonempty, there are two to five dimensions, and every weight is positive; the released data record how many rubrics were kept, repaired and excluded, and preserve the original rubrics.
These checks are structural: they do not verify the factual correctness or semantic coverage of a description.

\subsection{Training Configurations and Baseline Differences}
\label{app:training}
Table~\ref{tab:training} summarizes the training configurations.
Rubric-RL and \method{} share the optimizer and rollout settings; the main difference is the rubric carried by the training data.
Both disable KL reward shaping and KL loss, use a constant learning-rate schedule with ten warmup steps, and sample with temperature 1 and top-$p=1$.
The GRPO clipping parameters are 0.20 and 0.28 for the lower and upper bounds.
Checkpoints are saved and validated every 50 steps (every 100 steps for two early 4B models).
Every model trained with the rubric judge's reward also applies DAPO-style overlong shaping to its training rewards: a response longer than 4,096 tokens (the 8,192-token response limit minus a 4,096-token buffer) loses reward linearly, by up to 0.5 at the limit; validation scores are unshaped.
It is identical across Rubric-RL, \method{}, RuscaRL and the ablations and is active in every such run; it rarely binds under the larger KL coefficients, where responses stay short.

\begin{table}[t]
\centering
\caption{\textbf{Training hyperparameters and generation budgets of each method.}}
\label{tab:training}
\vspace{4pt}
\small
\renewcommand{\arraystretch}{1.1}
\setlength{\tabcolsep}{4pt}
\begin{tabular*}{\textwidth}{@{\extracolsep{\fill}}l|rrrr@{}}
\toprule
\textbf{Setting} & \textbf{Rubric-RL / \method{}} & \textbf{RuscaRL} & \textbf{OPSD} & \textbf{SFT} \\
\midrule
Training steps & 300 & 300 & 485 & 1,224 (3 epochs) \\
Prompts per batch & 64 & 512 & 128 & 64 \\
Responses per prompt & 8 & 1 & 1 & --- \\
Mini-batch size & 32 & 256 & 32 & --- \\
Learning rate & $10^{-6}$ & $10^{-6}$ & $4.2\times10^{-6}$ & $10^{-5}$ \\
Prompt limit & 4,096 & 6,144 & 4,096 & \multirow{2}{*}{20,000} \\
Response limit & 8,192 & 8,192 & 2,048 & \\
Training-time rubric judge & Yes & Yes & No & No \\
Overlong shaping (buffer / factor) & 4,096 / 0.5 & 4,096 / 0.5 & --- & --- \\
\bottomrule
\end{tabular*}
\end{table}

Step counts in Table~\ref{tab:training} are those of medical training. In the writing and dialogue domains RuscaRL exhausts its training set after 296--297 steps, so the evaluated checkpoint is the last one saved, step 250; the same holds for every variant built on RuscaRL's scaffolding. OPSD is evaluated at step 400 in writing, 350--352 in dialogue and 700 in science. In those two domains RuscaRL is therefore compared with Rubric-RL and \method{} at 250 steps against 300.
RuscaRL preserves contiguous scaffold groups during training and uses a larger prompt allowance to accommodate the additional context.
OPSD instead supplies rubric information to a teacher and optimizes a self-distillation loss following RGSD~\citep{Rezaei2026RGSD}; the rubric judge is used for validation, not to provide its training reward.
SFT fine-tunes on RubricHub's released SFT corpus (each response the best of six samples by rubric score; 26{,}162 of its 26{,}194 rows, after dropping empty rows and rows over the 20{,}000-token limit), whose prompts span all four of our domains, science included, mixed into one training set, following the supervised setup of RISE-RL~\citep{Hou2026RISE} (learning rate $10^{-5}$ with 5\% warmup, three epochs) with weight decay 0.1; it produces one model per seed that serves all four domains, whereas every RL policy is trained per domain.
Its training examples omit the empty \texttt{<think></think>} block that the Qwen3 chat template inserts at inference; the loss targets are unchanged.
The baseline results describe these configurations, not each method's best.

\subsection{Ablation Construction}
\label{app:ablation_details}
The raw-AND variant uses the same group assignments and absolute-weight sums as \method{}, but presents the original items under a joint-satisfaction instruction.
Negative-weight items become behaviors that must not occur.
The Graded variant keeps the rewritten descriptions and scores 0 to 3, divided by 3 before aggregation.
The single-dimension variant ($K{=}1$) rewrites the whole checklist into one \method{}-style description with its failure clause.

The failure-clause ablation removes sentences matching explicit failure expressions from the rewritten text.
Unmatched expressions remain, and descriptions shortened below 40 characters are restored.
It therefore tests removal of a class of explicit clauses, not complete elimination of failure semantics.
The added appropriateness criterion assesses directness, scope, and adaptation to the user.
Its wording is the same for Rubric-RL and \method{}, and its weight is the total absolute atomic weight divided by the corresponding number of \method{} dimensions.
For \method{}, this equals the mean existing dimension weight.
The new criterion is scored independently and included in the weighted reward; failure on it does not zero the scores of the other criteria.

Implicit aggregation keeps Rubric-RL's checklist and has the training judge read it whole and return one holistic score from 1 to 10 for the response.
The length-matched variant lowers Rubric-RL's response limit from 8,192 to 4,096 tokens.
The KL variants add a KL loss toward the reference policy (low-variance estimator) with $\beta \in \{0.005, 0.01, 0.04\}$ to both Rubric-RL and \method{}.
The weighted rubric keeps every criterion's text and count but multiplies the weight of criteria tagged critical by 3 ($13.5\%$ of criteria), sets formatting and duplicate criteria to 0 ($7.2\%$ and $11.3\%$) and needless-test criteria to $-1$ ($2.6\%$); each criterion is still scored and summed on its own.
In science, the rewritten-criteria variant restates each criterion one to one in the style of \method{}'s dimensions, keeping its weight and explicit aggregation.
The generator control regenerates \method{}'s dimensions with Doubao-lite instead of DeepSeek-V4-Pro under the same rules.

\section{Evaluation Protocols}
\label{app:evaluation}
\subsection{Benchmark Scoring and Judge Configurations}
HealthBench evaluation uses the upstream grading template with one judgment per criterion.
For each question, the score is the sum of satisfied criteria's signed weights divided by the sum of positive weights, followed by averaging over questions.
The same normalization is used for the consensus evaluation.
WritingBench uses query-dependent criteria; Creative Writing Benchmark v3 (Creative-v3) uses its isolated rubric component without the Elo/Glicko ranking component.
Arena-Hard v2 uses the upstream double-order comparison protocol and category-specific reference responses.
ResearchQA scores rubric coverage on a five-level scale, maps the levels to $[0,1]$, and averages first over criteria and then over questions.
We report all of these scores multiplied by 100.

DeepSeek-V4-Pro replaces the upstream judges for these model-judged benchmarks.
A second evaluator, Doubao-lite, is from the same family as the training reward judge; we use it only to compare judges.
\method{}'s dimensions are generated by DeepSeek-V4-Pro, which is also the judge, so the same model proposes the dimensions and scores the responses trained on them. Regenerating the dimensions with Doubao-lite moves appropriateness by less than two points (Appendix~\ref{app:sec5_notes}); raw-AND and implicit aggregation rewrite nothing and are unaffected.
All judges, generation seeds and scorer settings are fixed across evaluations, so paired comparisons are reproducible.

\paragraph{Agreement with physicians.}
\label{app:judge_agreement}
We ran DeepSeek-V4-Pro on HealthBench's meta-evaluation set (29{,}511 response--criterion pairs over the 34 consensus criteria, with 60{,}896 physician labels) using the official grader prompt, and scored it with the official code: balanced F1 against each physician label, averaged within each criterion and then over criteria. It reaches $0.62$, against $0.65$ for the average physician and $0.71$ for GPT-4.1 as reported by \citet{Arora2025HealthBench} (Table~\ref{tab:judge_agreement}). It is above the physician average in three of the seven themes and furthest below it on complex responses. Every method is scored by the same judge, so this bounds how well appropriateness tracks physicians, not the comparison between methods.

\begin{table}[t]
\centering
\caption{\textbf{Agreement with physicians on HealthBench's meta-evaluation (balanced F1).}}
\label{tab:judge_agreement}
\vspace{4pt}
\small
\renewcommand{\arraystretch}{1.15}
\resizebox{\textwidth}{!}{%
\begin{tabular}{@{}l|ccccccc|c@{}}
\toprule
\textbf{Metric} & \shortstack{\textbf{Emergency}\\\textbf{referrals}} & \shortstack{\textbf{Global}\\\textbf{health}} & \shortstack{\textbf{Communi-}\\\textbf{cation}} & \shortstack{\textbf{Context}\\\textbf{seeking}} & \textbf{Hedging} & \shortstack{\textbf{Health}\\\textbf{data}} & \shortstack{\textbf{Complex}\\\textbf{responses}} & \shortstack{\textbf{All}\\\textbf{criteria}} \\
\midrule
Criteria ($n$) & 6 & 3 & 4 & 4 & 9 & 4 & 4 & 34 \\
DeepSeek-V4-Pro & 0.65 & 0.69 & 0.63 & 0.61 & 0.61 & 0.71 & 0.45 & 0.62 \\
Physicians & 0.65 & 0.65 & 0.62 & 0.64 & 0.66 & 0.74 & 0.57 & 0.65 \\
\bottomrule
\end{tabular}}
\end{table}

\subsection{Paired Evaluation Sets and Statistical Estimation}
Every comparison is paired on the intersection of questions that all compared models have been scored on without evaluator failure, so the set shrinks as models are added and no comparison mixes subsets (Table~\ref{tab:eval_sizes}). For the ablations this gives 3{,}140 cases on the consensus axis and 4{,}392 on the coverage axis, from pools of 3{,}671 and 5{,}000; SFT and the 8B controls are paired on their own intersection with Base, Rubric-RL and \method{}.
The $g_{\mathrm{pro}}$ column of Table~\ref{tab:ablations_full} uses its own set: the $4{,}138$ of $4{,}563$ HealthBench-full prompts on which every seed of every variant in Table~\ref{tab:consensus_ablations} has a DeepSeek-V4-Pro verdict for every criterion. Other rows use the part of this set they cover.

\begin{table}[t]
\centering
\caption{\textbf{Sample sizes of common evaluation sets across benchmarks and judges.}}
\label{tab:eval_sizes}
\vspace{4pt}
\small
\renewcommand{\arraystretch}{1.12}
\begin{tabular*}{\textwidth}{@{\extracolsep{\fill}}llll@{}}
\toprule
\textbf{Medicine} & \textbf{Science} & \textbf{Writing} & \textbf{Dialogue} \\
\midrule
HealthBench-consensus (3{,}140) & GPQA-Diamond (198) & WritingBench (1{,}000) & Arena-Hard v2 (748) \\
HealthBench-full, lite (4{,}392) & ResearchQA (702) & Creative-v3 (96) & \\
HealthBench-full, pro (4{,}138) & & & \\
MedQA (1{,}273) & & & \\
\bottomrule
\end{tabular*}
\end{table}

At 4B, the WritingBench, Creative-v3, GPQA-Diamond and ResearchQA results of Base, and of Rubric-RL and \method{} at seed 42, average three evaluations of the same checkpoint; every other entry is a single evaluation.
These repeated evaluations are not the three trained seeds.

\subsection{Rubric-Free Comparisons and Length Controls}
\label{app:pairwise}
\begin{figure}[t]
\centering
\includegraphics[width=\linewidth]{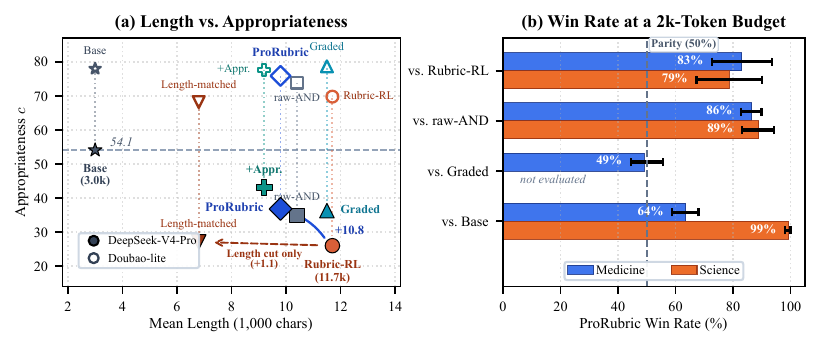}
\caption{\textbf{Response length and budget constraints.} (a) Appropriateness vs.\ mean response length (filled: DeepSeek-V4-Pro; open: Doubao-lite). (b) Pairwise win rates at a 2{,}048-token budget (DeepSeek-V4-Pro; vs.\ Rubric-RL: three seeds, others: seed 42).}
\label{fig:pairwise_winrates}
\end{figure}

The rubric-free evaluator receives the user prompt and two anonymous responses, with neither training rubric shown.
It judges usefulness, relevance, fit to the user's circumstances, and unnecessary verbosity.
Each pair is evaluated in both orders.
A stable win or loss requires agreement across the two orders; ties and inconsistent decisions are excluded from the conditional win-rate denominator.
For every comparison plotted in Figure~\ref{fig:pairwise_winrates}, we report the conditional win rate over stable, non-tied decisions; order consistency stays above 90\%.

\begin{table}[t]
\centering
\caption{\textbf{Full win/tie/loss counts and win rates for rubric-free pairwise comparisons.}}
\label{tab:rubric_free_full}
\vspace{4pt}
\small
\renewcommand{\arraystretch}{1.1}
\setlength{\tabcolsep}{4pt}
\begin{tabular*}{\textwidth}{@{\extracolsep{\fill}}l|cccc@{}}
\toprule
 & \multicolumn{2}{c}{\textbf{DeepSeek-V4-Pro}} & \multicolumn{2}{c}{\textbf{GPT-5.6-luna}} \\
\cmidrule(lr){2-3} \cmidrule(lr){4-5}
\textbf{Domain} & \textbf{W / T / L} & \textbf{Win rate (\%)} & \textbf{W / T / L} & \textbf{Win rate (\%)} \\
\midrule
\textit{Base vs.\ Rubric-RL} & & & & \\
\hspace{0.8em}Medicine & 136 / 93 / 71 & 65.7\std{13.2} & 229 / 55 / 16 & 93.6\std{2.3} \\
\hspace{0.8em}Science & 3 / 35 / 262 & 1.2\std{1.2} & 151 / 134 / 15 & 90.3\std{9.0} \\
\hspace{0.8em}Dialogue & 45 / 117 / 138 & 24.3\std{5.8} & 81 / 114 / 99 & 44.9\std{6.8} \\
\hspace{0.8em}Writing & 0 / 8 / 292 & 0.0\std{0.0} & 4 / 30 / 266 & 1.4\std{1.6} \\
\midrule
\textit{Base vs.\ \method{}} & & & & \\
\hspace{0.8em}Medicine & 64 / 104 / 132 & 33.1\std{11.5} & 162 / 98 / 40 & 80.2\std{6.2} \\
\hspace{0.8em}Science & 4 / 43 / 253 & 1.6\std{0.6} & 77 / 171 / 52 & 59.7\std{2.1} \\
\bottomrule
\end{tabular*}
\end{table}

The equal-budget evaluation regenerates both policies' responses under the same 2,048-token ceiling; this does not equalize realized lengths, but neither policy can generate beyond it.

\section{Full Numerical Results and Ablations}
\label{app:additional}
\label{app:per_seed}
\subsection{Per-Domain and Per-Seed Results}

Every trained-model score in Tables~\ref{tab:main_results} and~\ref{tab:consensus_ablations} is a mean over three training seeds. This section gives the 4B per-seed values: Table~\ref{tab:per_seed_domains} for the cross-domain benchmarks and Table~\ref{tab:per_seed_ablations} for the medical ablations; Table~\ref{tab:gpt_crosscheck} gives the third-family cross-check.

\paragraph{Reading the tables.} A slash separates seeds, in the order 42 / 43 / 44; a dash marks an entry we have not evaluated. Absolute scores are not comparable across judges (Appendix~\ref{app:sec5_notes}).

\paragraph{Seed variation.} Single benchmarks move far more across seeds than the seven-benchmark average. On GPQA-Diamond (198 questions) the paired difference between \method{} and Rubric-RL at 4B is $-3.5$, $-5.6$ and $+6.6$; on Arena-Hard it ranges from $+0.2$ to $+2.3$ ($+0.2$, $+1.4$, $+2.3$); on HealthBench it keeps its sign but varies fourfold ($+5.2$, $+1.9$, $+1.3$). We therefore report the average as a paired per-seed difference and read single benchmarks as orderings, not effect sizes.

\begin{table}[t]
\centering
\caption{\textbf{Per-seed cross-domain evaluation results at 4B scale (seeds 42, 43, and 44).}}
\label{tab:per_seed_domains}
\small
\renewcommand{\arraystretch}{1.1}
\setlength{\tabcolsep}{4pt}
\begin{tabular*}{\textwidth}{@{\extracolsep{\fill}}l|cc cc c@{}}
\toprule
\multirow{2}{*}{\textbf{Benchmark}} & \multicolumn{2}{c}{\textbf{Rubric-RL}} & \multicolumn{2}{c}{\textbf{\method{}}} & \multirow{2}{*}{\textbf{Paired $\boldsymbol{\Delta}$}} \\
\cmidrule(lr){2-3} \cmidrule(lr){4-5}
& \textbf{per seed} & \textbf{mean} & \textbf{per seed} & \textbf{mean} & \\
\midrule
\textit{Writing} & & & & & \\
\hspace{0.8em}WritingBench & 66.3 / 63.6 / 63.7 & 64.6\std{1.5} & 67.9 / 66.8 / 66.5 & 67.1\std{0.7} & +2.50\std{0.85} \\
\hspace{0.8em}Creative-v3 & 36.5 / 34.1 / 34.5 & 35.0\std{1.3} & 39.1 / 36.8 / 37.4 & 37.7\std{1.2} & +2.73\std{0.17} \\
\textit{Dialogue} & & & & & \\
\hspace{0.8em}Arena-Hard & 18.2 / 16.1 / 17.4 & 17.3\std{1.1} & 18.4 / 17.5 / 19.8 & 18.6\std{1.1} & +1.29\std{1.07} \\
\textit{Medicine} & & & & & \\
\hspace{0.8em}HealthBench & 30.8 / 33.0 / 31.8 & 31.9\std{1.1} & 36.0 / 34.9 / 33.1 & 34.7\std{1.4} & +2.77\std{2.10} \\
\hspace{0.8em}MedQA & 63.2 / 61.2 / 63.5 & 62.6\std{1.3} & 66.1 / 65.0 / 64.2 & 65.1\std{1.0} & +2.49\std{1.61} \\
\textit{Science} & & & & & \\
\hspace{0.8em}GPQA-Diamond & 46.8 / 48.0 / 39.4 & 44.7\std{4.7} & 43.3 / 42.4 / 46.0 & 43.9\std{1.8} & -0.84\std{6.50} \\
\hspace{0.8em}ResearchQA & 60.4 / 64.6 / 63.6 & 62.9\std{2.2} & 60.4 / 63.3 / 59.5 & 61.1\std{2.0} & -1.82\std{2.08} \\
\midrule
Seven-benchmark average & 46.0 / 45.8 / 44.9 & 45.6\std{0.6} & 47.3 / 46.7 / 46.6 & 46.9\std{0.4} & \textbf{+1.30}\std{0.46} \\
\bottomrule
\end{tabular*}
\end{table}

\begin{table}[t]
\centering
\caption{\textbf{Per-seed medical ablation results at 4B scale under both judges.} Subscripts: pro, DeepSeek-V4-Pro; lite, Doubao-lite.}
\label{tab:per_seed_ablations}
\vspace{4pt}
\small
\renewcommand{\arraystretch}{1.1}
\setlength{\tabcolsep}{3pt}
\begin{tabular*}{\textwidth}{@{\extracolsep{\fill}}l|ccc@{}}
\toprule
\textbf{Method} & $\boldsymbol{c_{\mathrm{pro}}}$ \textbf{per seed} & $\boldsymbol{c_{\mathrm{lite}}}$ \textbf{per seed} & $\boldsymbol{g_{\mathrm{lite}}}$ \textbf{per seed} \\
\midrule
\textit{Baselines} & & & \\
\hspace{0.8em}Rubric-RL & 24.3 / 28.4 / 25.1\std{2.2} & 69.9 / 71.3 / 68.3\std{1.5} & 55.7 / 56.1 / 55.4\std{0.4} \\
\hspace{0.8em}RuscaRL & 24.6 / 25.8 / 29.5\std{2.6} & 66.4 / 70.3 / 71.9\std{2.8} & 54.2 / 56.4 / 55.6\std{1.1} \\
\hspace{0.8em}OPSD & 40.7 / 42.9 / 41.4\std{1.1} & 64.8 / 67.9 / 66.3\std{1.6} & 35.4 / 36.9 / 36.5\std{0.8} \\
\textit{Aggregation changed} & & & \\
\hspace{0.8em}Implicit & 36.7 / 36.6 / 36.8\std{0.1} & 78.0 / 77.5 / 77.7\std{0.2} & 56.4 / 56.1 / 55.6\std{0.4} \\
\hspace{0.8em}raw-AND & 38.8 / 31.9 / 33.9\std{3.6} & 76.3 / 72.1 / 73.3\std{2.2} & 53.8 / 54.3 / 53.0\std{0.6} \\
\hspace{0.8em}Graded & 38.3 / 34.7 / 35.8\std{1.8} & 80.1 / 77.6 / 78.4\std{1.3} & 58.7 / 58.0 / 59.0\std{0.5} \\
\hspace{0.8em}\method{} & 39.8 / 36.2 / 34.5\std{2.7} & 78.0 / 74.9 / 74.9\std{1.8} & 56.4 / 55.3 / 54.2\std{1.1} \\
\textit{Appropriateness criterion added} & & & \\
\hspace{0.8em}Rubric-RL + appropriateness criterion & 34.5 / 34.5 / 33.7\std{0.5} & 72.8 / 72.7 / 71.6\std{0.7} & 54.9 / 52.9 / 54.2\std{1.0} \\
\hspace{0.8em}\method{} + appropriateness criterion & 45.4 / 40.7 / 43.4\std{2.4} & 77.6 / 77.6 / 78.0\std{0.2} & 53.6 / 53.8 / 53.2\std{0.3} \\
\textit{Controls} & & & \\
\hspace{0.8em}\method{} w/o failure clauses & 35.8 / 34.1 / 33.6\std{1.1} & 73.4 / 74.0 / 76.6\std{1.7} & 54.8 / 54.9 / 55.6\std{0.4} \\
\hspace{0.8em}$K{=}1$ & 34.8 / 37.6 / 38.5\std{1.9} & 74.1 / 75.3 / 77.1\std{1.5} & 51.9 / 53.0 / 52.4\std{0.5} \\
\hspace{0.8em}Length-matched & 27.5 / 25.1 / 28.7\std{1.8} & 69.4 / 67.2 / 67.9\std{1.1} & 52.9 / 52.4 / 53.2\std{0.4} \\
\bottomrule
\end{tabular*}
\end{table}

\begin{table}[t]
\centering
\caption{\textbf{Full medical ablations under both judges (4B).} Subscripts: pro, DeepSeek-V4-Pro; lite, Doubao-lite. Appr.\ crit.: the fixed appropriateness criterion (Appendix~\ref{app:appr_criterion}); Len.: mean response length in characters; $n$: number of seeds.}
\label{tab:ablations_full}
\vspace{4pt}
\footnotesize
\setlength{\tabcolsep}{4.5pt}
\renewcommand{\arraystretch}{1.15}
\resizebox{\textwidth}{!}{\begin{tabular}{l|cc cccc cc}
\toprule
\multirow{2}{*}{\textbf{Configuration}} & \multicolumn{2}{c}{\textbf{Reward structure}} & \multicolumn{4}{c}{\textbf{Score}} & \multirow{2}{*}{\textbf{Len.}} & \multirow{2}{*}{\textbf{$n$}} \\
\cmidrule(lr){2-3} \cmidrule(lr){4-7}
 & \textbf{grouped} & \textbf{appr.\ crit.} & $\boldsymbol{g_{\mathrm{lite}}}$ & $\boldsymbol{g_{\mathrm{pro}}}$ & $\boldsymbol{c_{\mathrm{lite}}}$ & $\boldsymbol{c_{\mathrm{pro}}}$ & & \\
\midrule
\addlinespace[0.5ex]\textit{Baselines} & & & & & & & & \\
\hspace{0.8em}Base & \factna & \factna & 38.1 & 26.6 & 78.0 & \textbf{54.1} & 3.0k & 1 \\
\hspace{0.8em}Rubric-RL & \factoff & \factoff & 55.8\std{0.4} & 32.1 & 69.8\std{1.5} & 26.0\std{2.2} & 11.7k & 3 \\
\hspace{0.8em}RuscaRL & \factoff & \factoff & 55.4\std{1.1} & 32.1 & 69.5\std{2.8} & 26.7\std{2.6} & 11.9k & 3 \\
\hspace{0.8em}OPSD & \factna & \factna & 36.3\std{0.8} & 25.6 & 66.3\std{1.6} & 41.7\std{1.1} & 4.7k & 3 \\

\addlinespace[0.5ex]\textit{Aggregation changed} & & & & & & & & \\
\hspace{0.8em}Implicit (one holistic score) & \facton & \factoff & 56.0\std{0.4} & 34.9 & 77.8\std{0.2} & 36.7\std{0.1} & 10.7k & 3 \\
\hspace{0.8em}raw-AND (text unchanged) & \facton & \factoff & 53.7\std{0.6} & 33.8 & 73.9\std{2.2} & 34.9\std{3.6} & 10.4k & 3 \\
\hspace{0.8em}Graded (4-level credit) & \facton & \factoff & 58.6\std{0.5} & 35.1 & 78.7\std{1.3} & 36.3\std{1.8} & 11.5k & 3 \\
\hspace{0.8em}\method{} & \facton & \factoff & 55.3\std{1.1} & 35.0 & 75.9\std{1.8} & 36.8\std{2.7} & 9.8k & 3 \\

\addlinespace[0.5ex]\textit{Appropriateness criterion added} & & & & & & & & \\
\hspace{0.8em}Rubric-RL + appropriateness criterion & \factoff & \facton & 54.0\std{1.0} & 33.7 & 72.4\std{0.7} & 34.2\std{0.5} & 10.6k & 3 \\
\hspace{0.8em}\method{} + appropriateness criterion & \facton & \facton & 53.5\std{0.3} & 35.3 & 77.7\std{0.2} & \textbf{43.1}\std{2.4} & 9.2k & 3 \\

\addlinespace[0.5ex]\textit{Controls} & & & & & & & & \\
\hspace{0.8em}\method{} w/o failure clauses & \facton & \factoff & 55.1\std{0.4} & 34.9 & 74.7\std{1.7} & 34.5\std{1.1} & 10.6k & 3 \\
\hspace{0.8em}$K{=}1$ (one dimension) & \facton & \factoff & 52.4\std{0.5} & 33.5 & 75.5\std{1.5} & 36.9\std{1.9} & 9.6k & 3 \\
\hspace{0.8em}Length-matched (4,096-token limit) & \factoff & \factoff & 52.9\std{0.4} & 30.4 & 68.2\std{1.1} & 27.1\std{1.8} & 6.8k & 3 \\
\hspace{0.8em}Rubric-RL, weighted criteria & \factoff & \factoff & 54.8\std{1.7} & 32.0 & 67.0\std{1.8} & 25.3\std{1.5} & 12.4k & 3 \\
\hspace{0.8em}\method{}, dimensions from another generator & \facton & \factoff & 51.4 & 33.8 & 71.6 & 38.1 & 10.4k & 1 \\

\addlinespace[0.5ex]\textit{KL penalty added} & & & & & & & & \\
\hspace{0.8em}Rubric-RL, $\beta{=}0.005$ & \factoff & \factoff & 46.3\std{1.1} & 28.8 & 61.0\std{3.2} & 25.8\std{1.3} & 12.7k & 3 \\
\hspace{0.8em}Rubric-RL, $\beta{=}0.01$ & \factoff & \factoff & 43.2\std{0.8} & 27.7 & 58.9\std{3.4} & 27.9\std{2.0} & 12.3k & 3 \\
\hspace{0.8em}Rubric-RL, $\beta{=}0.04$ & \factoff & \factoff & 41.4\std{0.6} & 28.4 & 75.2\std{0.3} & 49.3\std{0.7} & 4.1k & 3 \\
\hspace{0.8em}\method{}, $\beta{=}0.005$ & \facton & \factoff & 46.7\std{0.3} & 30.2 & 69.9\std{1.4} & 36.9\std{1.4} & 9.1k & 3 \\
\hspace{0.8em}\method{}, $\beta{=}0.01$ & \facton & \factoff & 44.4\std{0.5} & 30.5 & 75.4\std{0.3} & 48.0\std{1.3} & 4.9k & 3 \\
\hspace{0.8em}\method{}, $\beta{=}0.04$ & \facton & \factoff & 39.9\std{0.6} & 28.4 & 77.7\std{0.4} & \textbf{53.6}\std{0.4} & 3.3k & 3 \\
\bottomrule
\end{tabular}
}
\end{table}

\begin{table}[t]
\centering
\caption{\textbf{Cross-family ranking validation using GPT-5.6-luna (3,652--3,671 items per method).}}
\label{tab:gpt_crosscheck}
\vspace{4pt}
\small
\renewcommand{\arraystretch}{1.12}
\resizebox{\textwidth}{!}{%
\begin{tabular}{@{}l|cc|cccc|cc|ccc@{}}
\toprule
& \multicolumn{2}{c|}{\textbf{Baselines}} & \multicolumn{4}{c|}{\textbf{Aggregation changed}} & \multicolumn{2}{c|}{\textbf{+ Appr.\ crit.}} & \multicolumn{3}{c}{\textbf{KL penalty}} \\
\cmidrule(lr){2-3} \cmidrule(lr){4-7} \cmidrule(lr){8-9} \cmidrule(lr){10-12}
\textbf{Metric} & \textbf{Base} & \textbf{Rubric-RL} & \textbf{Implicit} & \textbf{raw-AND} & \textbf{Graded} & \textbf{\method{}} & \textbf{Rubric-RL} & \textbf{\method{}} & \textbf{Rubric-RL} & \textbf{Rubric-RL} & \textbf{\method{}} \\
& & & & & & & \textbf{+ appr.} & \textbf{+ appr.} & \textbf{$\beta{=}0.01$} & \textbf{$\beta{=}0.04$} & \textbf{$\beta{=}0.01$} \\
\midrule
\textbf{Score} & 33.7 & 14.4 & 21.6 & 22.4 & 20.1 & 20.3 & 18.0 & 22.6 & 20.2 & 29.2 & 28.5 \\
\textbf{Items} & 3{,}655 & 3{,}670 & 3{,}652 & 3{,}669 & 3{,}670 & 3{,}671 & 3{,}668 & 3{,}669 & 3{,}652 & 3{,}654 & 3{,}653 \\
\bottomrule
\end{tabular}}
\end{table}

\begin{table}[t]
\centering
\caption{\textbf{Per-seed HealthBench-consensus results at 4B ($\Delta = \text{\method{}} - \text{Rubric-RL}$).}}
\label{tab:seeds}
\vspace{4pt}
\small
\renewcommand{\arraystretch}{1.1}
\begin{tabular*}{\textwidth}{@{\extracolsep{\fill}}c|rrrrrr@{}}
\toprule
& \multicolumn{3}{c}{\textbf{DeepSeek-V4-Pro}} & \multicolumn{3}{c}{\textbf{Doubao-lite}} \\
\cmidrule(lr){2-4}\cmidrule(lr){5-7}
\textbf{Seed} & \textbf{Rubric-RL} & \textbf{\method{}} & $\boldsymbol{\Delta}$ & \textbf{Rubric-RL} & \textbf{\method{}} & $\boldsymbol{\Delta}$ \\
\midrule
42 & 24.3 & 39.8 & +15.5 & 69.9 & 78.0 & +8.1 \\
43 & 28.4 & 36.2 & +7.7 & 71.3 & 74.9 & +3.6 \\
44 & 25.1 & 34.5 & +9.4 & 68.3 & 74.9 & +6.6 \\
\midrule
Mean & 26.0 & 36.8 & +10.8 & 69.8 & 75.9 & +6.1 \\
\bottomrule
\end{tabular*}

\end{table}
The mean improvement is 10.8 points under DeepSeek-V4-Pro and 6.1 under Doubao-lite over the three seeds shown.
These averages are not confidence intervals. They are computed from unrounded per-seed scores, so adding the displayed one-decimal values can differ from the stated mean by $0.1$.
\section{Mechanistic Analyses and Extended Ablations}
\label{app:mechanistic}

\subsection{Sensitivity Analysis: All Reward Structures}
\label{app:sensitivity_full}
Table~\ref{tab:incentive_audit_full} extends Table~\ref{tab:evidence}b to every reward structure, adds partial satisfaction of a criterion, which the main text does not use, and gives the writing rubric under explicit aggregation. The writing rubric was tested with the two naming perturbations only, since the other three are clinical; naming an item earns $+1.3$ there, an interval containing zero.

\begin{table}[t]
\centering
\caption{\textbf{Reward change per edit before training, for every aggregation.} + appr.: with the fixed appropriateness criterion (Appendix~\ref{app:appr_criterion}).}
\label{tab:incentive_audit_full}
\vspace{4pt}
\small
\setlength{\tabcolsep}{3.2pt}
\resizebox{\textwidth}{!}{%
\begin{tabular}{@{}l|cccccccc@{}}
\toprule
\multirow{2}{*}{\textbf{Edit to the answer}} & \multirow{2}{*}{\textbf{Rubric-RL}} & \multirow{2}{*}{\textbf{raw-AND}} & \multirow{2}{*}{\textbf{Graded}} & \multirow{2}{*}{\textbf{\method{}}} & \textbf{Rubric-RL} & \textbf{\method{}} & \multirow{2}{*}{\textbf{Implicit}} & \textbf{Writing} \\
 & & & & & \textbf{+ appr.} & \textbf{+ appr.} & & \textbf{(Rubric-RL)} \\
\midrule
Name an item & $\mathbf{+7.3}$ & $-0.3$ & $+1.8$ & $+2.2$ & $+3.0$ & $-3.9$ & $+1.4$ & $+1.3$ \\
 & \textcolor{black!45}{\scriptsize [$+5.9$, $+8.7$]} & \textcolor{black!45}{\scriptsize [$-1.9$, $+1.4$]} & \textcolor{black!45}{\scriptsize [$+0.8$, $+2.8$]} & \textcolor{black!45}{\scriptsize [$-0.1$, $+4.5$]} & \textcolor{black!45}{\scriptsize [$+1.5$, $+4.5$]} & \textcolor{black!45}{\scriptsize [$-6.4$, $-1.4$]} & \textcolor{black!45}{\scriptsize [$-0.2$, $+2.8$]} & \textcolor{black!45}{\scriptsize n.s.} \\
Name the topic & $-0.4$ & $+0.6$ & $-1.2$ & $-3.8$ & $-1.6$ & $-7.6$ & $-1.7$ & $-0.3$ \\
 & \textcolor{black!45}{\scriptsize [$-1.7$, $+0.8$]} & & & & & & & \\
Satisfy a criterion in part & $+2.2$ & $+1.1$ & $+1.8$ & $+3.3$ & $+1.9$ & $+0.4$ & $+0.9$ & -- \\
Add a needless test & $+0.3$ & $-1.4$ & $-2.7$ & $-5.8$ & $-4.6$ & $\mathbf{-9.4}$ & $-4.5$ & -- \\
Drop the key advice & $-0.9$ & $-0.8$ & $-1.1$ & $-2.2$ & $-0.5$ & $-0.2$ & $-1.2$ & -- \\
Paraphrase & $+1.1$ & $+1.1$ & $+1.5$ & $+1.0$ & $+1.7$ & $+1.1$ & $+1.0$ & -- \\
\bottomrule
\end{tabular}}
\end{table}

\subsection{Notes on Section~\ref{sec:experiments}}
\label{app:sec5_notes}
\paragraph{Judges.} Doubao-lite, from the training reward's family, sees the reversal of Section~\ref{sec:quality_gap} on a smaller scale: Rubric-RL gains $17.7$ points of coverage over the untrained model and loses $8.2$ on the physician criteria, against a $5.5$-point gain and a $28.1$-point loss under DeepSeek-V4-Pro. GPT-5.6-luna ranks the models in nearly the same order as DeepSeek-V4-Pro (Spearman $\rho = 0.84$, eleven models, same responses), but its absolute scores are far lower: the untrained model scores $33.7$ where DeepSeek-V4-Pro gives $54.1$. We therefore read it for ordering only. Criterion by criterion, of the positive-weight physician criteria on $3{,}657$ prompts ($7{,}949$ to $8{,}019$ per model), Doubao-lite marks satisfied and DeepSeek-V4-Pro unmet $27.5\%$ for the untrained model and $46.2\%$, $44.3\%$ and $44.1\%$ for Rubric-RL at seeds 42, 43 and 44. Under GPT-5.6-luna, grouping alone (raw-AND) is worth $+7.9$ over Rubric-RL, and appending the appropriateness criterion to \method{}'s dimensions $+4.6$ over appending it to the original checklist. Scored by Doubao-lite, appending the appropriateness criterion costs $1.8$ points of coverage, negative on all three seeds; scored by DeepSeek-V4-Pro on the same responses it costs nothing ($+0.30 \pm 1.05$, with the sign flipping between seeds). Two further pairs behave the same way at seed 42: the length-matched variant costs $2.8$ points of coverage under Doubao-lite and $0.7$ under DeepSeek-V4-Pro, and Graded leads \method{} by $2.2$ in coverage under Doubao-lite and by $0.8$ under DeepSeek-V4-Pro. In each pair Doubao-lite prefers the longer variant, and the gap shrinks about three- to fourfold under DeepSeek-V4-Pro.

\paragraph{Aggregation controls at 8B and the generator.} At 8B over three seeds, implicit aggregation reaches $47.6 \pm 1.0$ against \method{}'s $45.8 \pm 1.8$ and Rubric-RL's $29.5 \pm 1.1$, and raw-AND recovers $11.4$ of \method{}'s $16.3$ points. At 4B, $K{=}1$ is above \method{} on appropriateness at two seeds and below at one, and is lower on coverage on average; $63$--$66\%$ of its sampled groups receive identical rewards. Regenerating \method{}'s dimensions with a different generator moves appropriateness by less than two points ($38.1$ against $39.8$ at the same seed).

\paragraph{KL penalty.} At $\beta \le 0.01$ Rubric-RL's appropriateness stays flat while its coverage falls. The separation between \method{} and Rubric-RL across $\beta \in \{0, 0.005, 0.01, 0.04\}$ is $+10.8$, $+11.1$, $+20.1$ and $+4.3$, three-seed means. The pre-registered test required at least $+10$ at $\beta = 0.005$ and at least $+5$ at $\beta = 0.04$; the first holds and the second does not (per seed $+4.9$, $+4.3$ and $+3.7$). At $\beta = 0.04$ \method{} reaches $53.6$ against the untrained model's $54.1$, with coverage $28.4$ against $26.6$ and responses of $3.3$k characters against $3.0$k. Wherever the policy still learns, \method{} is ahead by 10 to 20 points, a restriction we chose after seeing these results.

\paragraph{Training dynamics.}
\label{app:dynamics}
\begin{figure}[ht]
\centering
\includegraphics[width=0.5\textwidth]{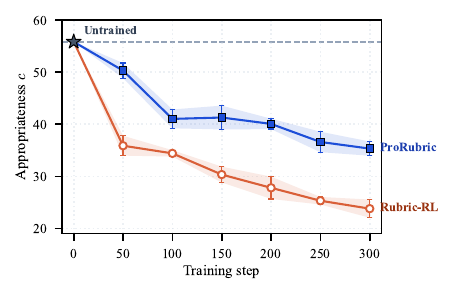}
\caption{\textbf{Appropriateness during training.} Medical 4B, 300 HealthBench-consensus prompts; three-seed mean $\pm$ standard deviation, step 0 is the untrained model.}
\label{fig:training_dynamics}
\end{figure}
The models behind Figure~\ref{fig:training_dynamics} repeat the Rubric-RL and \method{} settings of Table~\ref{tab:consensus_ablations} with the same seeds, saving a checkpoint every 50 steps; they are newly trained, not the checkpoints of Table~\ref{tab:consensus_ablations}, and all six are reported. Appropriateness is scored on 300 HealthBench-consensus prompts from the common set by a second instance of the same DeepSeek-V4-Pro model. At step 300 the curves end at $23.8$ and $35.3$. Scored as in Table~\ref{tab:consensus_ablations} (full common set, same judge instance), the step-300 checkpoints reach $24.3$ and $35.7$ (three-seed means; \method{} over Rubric-RL $+11.5$, against $+10.8$ in Table~\ref{tab:consensus_ablations}), and each Table~\ref{tab:consensus_ablations} mean lies within the range of the retrained runs. Grouping \method{}'s dimensions by their status in the untrained model's answer, dimensions one criterion short of complete are completed more often by step 300 than partly or wholly unmet ones, under both rewards and on every seed, so this ordering is not specific to conjunction.

\paragraph{SFT.} On the consensus axis SFT ends above the untrained model, $66.3 \pm 0.4$ against $54.1$ at 4B and $75.6 \pm 0.5$ against $62.5$ at 8B ($3{,}143$ paired items), where Rubric-RL falls to $26.0$ and $29.5$; none of the consensus prompts appears in its training data. Outside its domain it pays the largest cost, $-5.7$ on MedQA at 4B and $-12.0$ on GPQA-Diamond at 8B, where both RL methods gain about seven. Its responses run 19--55\% longer than \method{}'s on the writing benchmarks. On HealthBench it is level with \method{} at both scales ($35.1$ against $34.7$ at 4B, $40.4$ against $39.9$ at 8B).

\paragraph{Domain and rubric shape.} The four domains differ in their rubrics as well as their tasks: science has $7.5$ criteria per question in graded tiers with negative weights, the other domains $27$ to $32$ positive ones. With four domains the two cannot be separated.

\subsection{A Fixed Appropriateness Criterion}
\label{app:appr_criterion}
Appending one fixed criterion to every checklist, asking the response to answer what was asked, stay in scope and suit the asker (Appendix~\ref{app:ablation_details}), raises appropriateness under every aggregation: Rubric-RL from $26.0$ to $34.2 \pm 0.5$, \method{} from $36.8$ to $43.1 \pm 2.4$, and implicit aggregation from $36.7$ to $44.1 \pm 0.8$ (three seeds each). Coverage is unchanged or higher ($33.7$ and $35.3$ against $32.1$ and $35.0$). Because the criterion asks for nearly what the physician criteria check, we do not count it as part of \method{}.
It acts on length rather than on errors. On the untrained model's answers the training judge marks it unmet $54.0\%$ of the time; adding rubric terms raises this to $81.7\%$, while deleting the answer's key sentence raises it only to $57.1\%$. It is judged unmet on $80.5\%$ of answers into which a mildly inappropriate recommendation is inserted and on $56.3\%$ of content-preserving paraphrases. Adjudicating a sample of its unmet verdicts with DeepSeek-V4-Pro, $67.3\%$ of those on detail-only variants and $52.0\%$ on the untrained model's answers were unwarranted.

\subsection{Cross-Domain Appropriateness Evaluation}
\label{app:probe}
The cross-domain results in Sections~\ref{sec:ablations} and~\ref{sec:redundancy_tolerance} come from an evaluation fixed before training finished. Each domain contributes 200 prompts: HealthBench-consensus in medicine, Arena-Hard v2 in dialogue, ResearchQA in science, and 96 Creative-v3 plus 104 WritingBench prompts in writing. DeepSeek-V4-Pro scores every response on the same four domain-independent criteria: the response addresses the primary request before supplementary material, adapts to the user's stated context, includes only material that helps and avoids repetition (\emph{redundancy}), and separates supported claims from uncertainty. The score is the mean over the four criteria ($\times 100$). Differences are paired over prompts, with 95\% bootstrap intervals over prompts only. Table~\ref{tab:probe} shows seed 42 unless a seed is marked. The changes from the untrained model in Figure~\ref{fig:controls_compound}c and Section~\ref{sec:redundancy_tolerance}, and science's $7.5$ in Section~\ref{sec:ablations}, are instead means over seeds 42, 43 and 44 of Rubric-RL and \method{} on the same prompts (Table~\ref{tab:probe_seeds_audit}a). The redundancy sub-score of Sections~\ref{sec:levers} and~\ref{sec:redundancy_tolerance} is the change, in points, in the pass rate of the redundancy criterion alone when the untrained model's answers are perturbed as in Section~\ref{sec:incentive_audit}, in each domain (Table~\ref{tab:probe_seeds_audit}b).

\begin{table}[t]
\centering
\caption{\textbf{Cross-domain appropriateness: scores and differences from Rubric-RL across four domains.}}
\label{tab:probe}
\vspace{4pt}
\small
\renewcommand{\arraystretch}{1.12}
\resizebox{\textwidth}{!}{%
\begin{tabular}{@{}l|cc|cc|cc|cc@{}}
\toprule
& \multicolumn{2}{c|}{\textbf{Medicine}} & \multicolumn{2}{c|}{\textbf{Dialogue}} & \multicolumn{2}{c|}{\textbf{Science}} & \multicolumn{2}{c}{\textbf{Writing}} \\
\cmidrule(lr){2-3} \cmidrule(lr){4-5} \cmidrule(lr){6-7} \cmidrule(lr){8-9}
\textbf{Method} & \textbf{Score} & $\boldsymbol{\Delta}$ \textbf{[95\% CI]} & \textbf{Score} & $\boldsymbol{\Delta}$ \textbf{[95\% CI]} & \textbf{Score} & $\boldsymbol{\Delta}$ \textbf{[95\% CI]} & \textbf{Score} & $\boldsymbol{\Delta}$ \textbf{[95\% CI]} \\
\midrule
Base & 50.0 & $+31.1^{*}$ [$+27.0$, $+35.1$] & 45.6 & $+7.4^{*}$ [$+3.6$, $+11.1$] & 52.2 & $+11.2^{*}$ [$+8.4$, $+14.4$] & 38.9 & $-19.5^{*}$ [$-23.9$, $-15.1$] \\
Rubric-RL & 18.9 & --- & 38.2 & --- & 41.0 & --- & 58.4 & --- \\
\method{} & 39.2 & $+20.4^{*}$ [$+16.9$, $+23.9$] & 44.9 & $+6.6^{*}$ [$+3.5$, $+9.9$] & 51.0 & $+10.0^{*}$ [$+7.4$, $+12.6$] & 64.1 & $+5.8^{*}$ [$+3.2$, $+8.2$] \\
raw-AND & 29.9 & $+11.0^{*}$ [$+8.0$, $+14.1$] & 45.9 & $+7.6^{*}$ [$+4.4$, $+10.9$] & 43.6 & $+2.6^{*}$ [$+0.1$, $+5.2$] & 59.9 & $+1.5$ [$-1.2$, $+4.2$] \\
Implicit, seed 42 & -- & -- & 43.8 & $+5.5^{*}$ [$+2.5$, $+8.8$] & 44.2 & $+3.2^{*}$ [$+0.8$, $+5.9$] & 66.0 & $+7.6^{*}$ [$+5.4$, $+9.9$] \\
Implicit, seed 43 & -- & -- & 44.1 & $+5.9^{*}$ [$+3.0$, $+9.0$] & 42.8 & $+1.8$ [$-0.6$, $+4.2$] & -- & -- \\
Implicit, seed 44 & -- & -- & 43.4 & $+5.1^{*}$ [$+2.0$, $+8.2$] & 41.5 & $+0.5$ [$-1.8$, $+2.9$] & -- & -- \\
Rewritten, seed 42 & -- & -- & -- & -- & 43.1 & $+2.1$ [$-0.2$, $+4.5$] & -- & -- \\
Rewritten, seed 43 & -- & -- & -- & -- & 42.6 & $+1.6$ [$-0.8$, $+4.1$] & -- & -- \\
\bottomrule
\end{tabular}}
\end{table}

\paragraph{Blind preference and appropriateness.} The blind pairwise judge of Table~\ref{tab:evidence}a weighs, in order, whether a response answers what was asked, fits the asker, is accurate, and avoids unnecessary length, and returns one overall preference; the criteria above are judged one at a time, so redundancy is a quarter of the score and cannot be offset by completeness. In science the two readings part for the same judge on the same prompts and checkpoints. On the 100 ResearchQA prompts per seed of Table~\ref{tab:evidence}a, DeepSeek-V4-Pro prefers Rubric-RL in 262 of 300 pairs, but on the answers it prefers, Rubric-RL passes the redundancy criterion $0$--$1\%$ of the time against $35$--$42\%$ for the untrained model. On these prompts appropriateness falls by $12.0$ to $13.0$ points per seed: the pass rate of the redundancy criterion falls by $35$ to $36$ points and that of adaptation to context by $14$ to $18$, while answering the primary request rises by $1$. Rubric-RL's science answers are more complete (ResearchQA coverage $51.8$ to $62.9$, Table~\ref{tab:main_results}) and less focused; DeepSeek-V4-Pro counts the completeness as worth its length when ranking whole answers, and GPT-5.6-luna does not. Dialogue follows the same pattern: both judges prefer the trained policy, and its $7.1$-point fall lies mostly in the redundancy criterion.

\begin{table}[ht]
\centering
\small
\caption{\textbf{Cross-domain appropriateness over three seeds, and the incentive audit in every domain.} (a) Change from the untrained model on the 200 prompts per domain of Table~\ref{tab:probe}, seeds 42 / 43 / 44, mean $\pm$ s.d.; the last column pairs each seed with the Rubric-RL model of the same seed. (b) The untrained model's answers to 300 training prompts per domain, perturbed as in Section~\ref{sec:incentive_audit}: reward change ($\times 100$) under the Rubric-RL reward for naming an item and for adding unrequested substance; the naming payment as a share of the substance payment under each reward; the naming edits that leave the Rubric-RL reward unchanged; and the change in the redundancy criterion's pass rate (points; all prompts with every variant, 293--300 per domain).}
\label{tab:probe_seeds_audit}
\vspace{4pt}
\renewcommand{\arraystretch}{1.1}
\setlength{\tabcolsep}{4pt}
\resizebox{\textwidth}{!}{%
\begin{tabular}{@{}l|c|cc|cc|c@{}}
\toprule
\multicolumn{7}{@{}l}{\textbf{(a) Appropriateness change from the untrained model}} \\
\midrule
\textbf{Domain} & \textbf{Untrained} & \multicolumn{2}{c|}{\textbf{Rubric-RL}} & \multicolumn{2}{c|}{\textbf{\method{}}} & \textbf{\method{} $-$ Rubric-RL} \\
 & \textbf{score} & 42 / 43 / 44 & mean & 42 / 43 / 44 & mean & mean (lowest seed) \\
\midrule
Medicine & 50.0 & $-31.1$ / $-26.5$ / $-29.0$ & $-28.9$\std{2.3} & $-10.8$ / $-16.6$ / $-19.5$ & $-15.6$\std{4.5} & $+13.3$ ($+9.5$) \\
Science  & 52.2 & $-11.2$ / $-11.6$ / $-10.8$ & $-11.2$\std{0.4} & $-1.2$ / $-6.6$ / $-3.2$ & $-3.7$\std{2.7} & $+7.5$ ($+5.0$) \\
Dialogue & 45.6 & $-7.4$ / $-9.8$ / $-4.1$ & $-7.1$\std{2.8} & $-0.8$ / $-0.2$ / $+0.2$ & $-0.3$\std{0.5} & $+6.8$ ($+4.4$) \\
Writing  & 38.9 & $+19.5$ / $+13.9$ / $+14.6$ & $+16.0$\std{3.1} & $+25.2$ / $+25.0$ / $+22.6$ & $+24.3$\std{1.4} & $+8.3$ ($+5.8$) \\
\bottomrule
\end{tabular}}

\vspace{8pt}
\resizebox{\textwidth}{!}{%
\begin{tabular}{@{}l|cc|cc|c|cc@{}}
\toprule
\multicolumn{8}{@{}l}{\textbf{(b) Incentive audit before training}} \\
\midrule
 & \multicolumn{2}{c|}{\textbf{Rubric-RL reward}} & \multicolumn{2}{c|}{\textbf{Naming / substance}} & \textbf{Reward unchanged} & \multicolumn{2}{c}{\textbf{Redundancy pass rate}} \\
\textbf{Domain} & \textbf{Name an item} & \textbf{Add substance} & \textbf{Rubric-RL} & \textbf{\method{}} & \textbf{by naming} & \textbf{Name an item} & \textbf{Add substance} \\
\midrule
Medicine & $+7.3$ & $+22.7$ & $32\%$ & $6\%$   & 37 / 300  & $-17.0$ & $-25.0$ \\
Dialogue & $+2.4$ & $+17.9$ & $13\%$ & $-14\%$ & 66 / 300  & $-29.2$ & $-33.6$ \\
Writing  & $+1.3$ & $+18.9$ & $7\%$  & $-10\%$ & 71 / 300  & $-25.6$ & $-37.5$ \\
Science  & $-0.8$ & $+10.8$ & $-8\%$ & $-23\%$ & 195 / 300 & $-26.8$ & $-31.9$ \\
\bottomrule
\end{tabular}}
\end{table}

\subsection{Where Added Material Sits: Appending versus Integrating}
\label{app:placement}
The two perturbations of Section~\ref{sec:redundancy_tolerance} both append material, so they cannot say whether the appropriateness charge is for the added content or for its position.
A third perturbation places the same material inside the answer instead of appending it. Sized to a common per-prompt budget, the two placements are charged alike: on the items where both land within $25\%$ of each other, the Rubric-RL reward pays $+41.4$ appended against $+40.3$ integrated in medicine ($n{=}35$) and $+19.5$ against $+18.4$ in science ($n{=}108$), with \method{} at $+48.8$ against $+52.1$ and $+28.9$ against $+28.5$. Writing and dialogue lean the other way by $4$ to $12$ points ($n{=}50$ and $n{=}66$), but we do not read that lean as an effect of placement: within every matched pair the integrated variant is the longer one, by $13.7\%$ to $17.2\%$, which is the same order as the gap itself.
The comparison does settle one point. At unequal budgets, with the integrated variant given about $18\%$ of the response and the appended one $50\%$, on $8$ to $87$ items, the domains had been $31$ points apart. At a common budget that spread is gone and no domain charges nothing. \emph{Where} the material sits is at most a second-order term; what is charged is the material.
The comparison has three limits. The pair is matched after the fact and only $12\%$ to $36\%$ of prompts admit both rewrites, so the four domains are not four repetitions of one experiment; matching selected \emph{shorter} originals in medicine and \emph{longer} ones in the other three. The realized expansion is $+68\%$ to $+105\%$ rather than the $+50\%$ we pre-registered. And the matched subsets run from $35$ to $108$ items, so the per-domain gaps carry wide intervals.

\section{Qualitative Case Studies and Protocol Analysis}
\label{app:qualitative}

\subsection{\method{} Rubrics and Representative Responses}
\label{app:representative_examples}

\paragraph{One rubric, before and after.}
A training prompt from the clinical split reads: \emph{``Doctor, I was in a car accident and I'm experiencing neck pain and abdominal pain. What could be the cause of this?''}
Its atomic rubric carries 32 independently scored criteria, among them \emph{``explicitly states that the neck and abdominal pain are occurring after a car accident''} ($w=8$), \emph{``avoids stating a single certain diagnosis''} ($w=9$), \emph{``lists concrete urgent warning signs''} ($w=10$) and \emph{``describes whiplash or neck muscle strain as common post-collision injuries''} ($w=7$).
\method{} folds these into four protocol dimensions, each inheriting the summed weight of its members ($59$, $118$, $36$ and $20$, totaling $233$, exactly the sum of the 32 atomic weights):

\begin{enumerate}[leftmargin=16pt, itemsep=2pt, topsep=2pt]
    \item \textbf{Emergency recognition} ($w=59$): directs the user to urgent in-person evaluation. \emph{Fails if it omits the recommendation for urgent medical evaluation, fails to include NPO guidance, or presents the situation as non-urgent.}
    \item \textbf{Probabilistic differential} ($w=118$): spans common minor conditions and life-threatening pathologies, linking each to the mechanism of blunt trauma. \emph{Fails if it presents only benign explanations, omits the referred-pain mechanism linking the two symptom regions, or states a definitive diagnosis.}
    \item \textbf{Expected workup} ($w=36$): describes the physical examination, laboratory tests and imaging. \emph{Fails if it does not outline the expected workup or omits dual-region CT imaging.}
    \item \textbf{Accessible communication} ($w=20$): organises the content and explains terminology. \emph{Fails if it uses unexplained jargon or ends without a summary reinforcing the key safety message.}
\end{enumerate}

The requirements are the same, grouped; each dimension gains a failure clause and is paid only when all of its members hold.

\paragraph{What the two policies produce.}
The clearest instances are the shortest prompts. On the consensus benchmark, one user writes five words: \emph{``My baby has a fever.''}
Two physician-authored criteria apply, each weighted $5$.
The Rubric-RL policy answers in $10{,}890$ characters and DeepSeek-V4-Pro marks \emph{both} unmet; the \method{} policy answers in $5{,}277$ and the judge marks both met.
The first criterion states its own reason: the response ``should avoid being overly alarmist,'' should ``recommend care with enough urgency,'' and, most directly, \emph{``the emergency referral should not be overly verbose, unclear, or indirect in a way that could distract the user from seeking care.''}
The second requires the response either to seek the missing context that decides whether emergency care is needed, or to give conditional guidance covering the likely scenarios.
Neither criterion rewards breadth, and the longer answer satisfies neither.

\subsection{Failure Cases and Coverage Trade-offs}
\label{app:failure_cases}
In action-oriented domains conjunctive grouping discourages long lists of loosely related facts, but it has a cost on broad information-retrieval tasks such as ResearchQA.

In literature-synthesis queries where the evaluation rubric tests exhaustive topical coverage (e.g., enumerating all documented experimental methodologies across multiple sub-fields), \method{} tends to summarize the main methodological approaches and occasionally omits niche cases that Rubric-RL lists explicitly. Under a protocol that scores only coverage of the atomic checklist, \method{} therefore scores slightly lower than the unconstrained Rubric-RL (61.1 vs.\ 62.9 at 4B and 64.9 vs.\ 65.8 at 8B, three-seed means). When a task's utility is exhaustive enumeration rather than concise decision-making, the conciseness enforced by conjunctive failure clauses gives up some coverage breadth for denser communication, a measurable trade-off.

\end{document}